\documentclass[10pt,twocolumn,letterpaper]{article}

\usepackage[pagenumbers]{cvpr} 

\usepackage{graphicx}
\usepackage{amsmath}
\usepackage{cite}
\usepackage{booktabs}
\usepackage{amsmath}
\usepackage{multirow}
\usepackage{makecell}
\usepackage{pifont}
\newcommand{\cmark}{\ding{51}}%
\usepackage{amssymb}
\usepackage{xcolor}
\usepackage[export]{adjustbox}
\usepackage{hhline}
\newcolumntype{P}[1]{>{\centering\arraybackslash}p{#1}}

\usepackage[pagebackref,breaklinks,colorlinks]{hyperref}

\usepackage[capitalize]{cleveref}
\crefname{section}{Sec.}{Secs.}
\Crefname{section}{Section}{Sections}
\Crefname{table}{Table}{Tables}
\crefname{table}{Tab.}{Tabs.}

\def\cvprPaperID{6858} 
\def\confName{CVPR}
\def\confYear{2024}

\begin{document}

\title{Excision And Recovery: Visual Defect Obfuscation Based \\Self-Supervised Anomaly Detection Strategy}

\author{YeongHyeon Park$^{1, 2}$ \qquad Sungho Kang$^{1}$ \qquad Myung Jin Kim$^{2}$ \\
Yeonho Lee$^{1}$ \qquad Hyeong Seok Kim$^{2}$ \qquad Juneho Yi$^{1}$~\thanks{Corresponding author: Juneho Yi.} \\
$^{1}$Department of Electrical and Computer Engineering, Sungkyunkwan University\\
$^{2}$SK Planet Co., Ltd.\\
{\tt\small \{yeonghyeon, myungjin, beman\}@sk.com, \{sungho369, tarazed, jhyi\}@skku.edu }
}
\maketitle

\begin{abstract}
Due to scarcity of anomaly situations in the early manufacturing stage, an unsupervised anomaly detection (UAD) approach is widely adopted which only uses normal samples for training.
This approach is based on the assumption that the trained UAD model will accurately reconstruct normal patterns but struggles with unseen anomalous patterns.
To enhance the UAD performance, reconstruction-by-inpainting based methods have recently been investigated, especially on the masking strategy of suspected defective regions.
However, there are still issues to overcome:
1) time-consuming inference due to multiple masking,
2) output inconsistency by random masking strategy, and
3) inaccurate reconstruction of normal patterns when the masked area is large.
Motivated by this, we propose a novel reconstruction-by-inpainting method, dubbed \textit{Excision And Recovery} (EAR), that features single deterministic masking based on the ImageNet pre-trained DINO-ViT and visual obfuscation for hint-providing.
Experimental results on the MVTec AD dataset show that deterministic masking by pre-trained attention effectively cuts out suspected defective regions and resolve the aforementioned issues 1 and 2.
Also, hint-providing by mosaicing proves to enhance the UAD performance than emptying those regions by binary masking, thereby overcomes issue 3.
Our approach achieves a high UAD performance without any change of the neural network structure.
Thus, we suggest that EAR be adopted in various manufacturing industries as a practically deployable solution.
\end{abstract}

\vspace{-0.4cm}
\section{Introduction}

In the manufacturing industry, ensuring product quality is of paramount importance, which can be automated by machine vision systems~\cite{Inspection_CIRP16,Inspection_Sruthy_AC19}. 
Machine vision systems for defective product detection can be implemented with machine learning or deep learning-based models. 
However, a significant challenge arises when confronted with the scarcity of anomaly situations, leading to an imbalanced dataset during the early stages of manufacturing. 
In such cases, training of an anomaly detection (AD) model under full supervision becomes practically unfeasible.

Recognizing this predicament, the manufacturing industry has increasingly turned to an unsupervised anomaly detection (UAD) approach. 
The data imbalance problem is eased simply by UAD because it only exploits prevalent normal samples for the training stage and does not require any defective samples.
The rationale behind this approach hinges on the idea that a well-trained UAD model excels in the accurate reconstruction of normal patterns but falters when trying to reconstruct unseen anomalous patterns.
We refer to this as \textit{contained generalization ability}~\cite{LAMP_Park_arXiv23}.

\begin{figure*}[t]
    \resizebox{\linewidth}{!}{%
        \includegraphics*[width=1.0\linewidth,trim={0.0cm 0.0cm 0.0cm 0.4cm},clip]{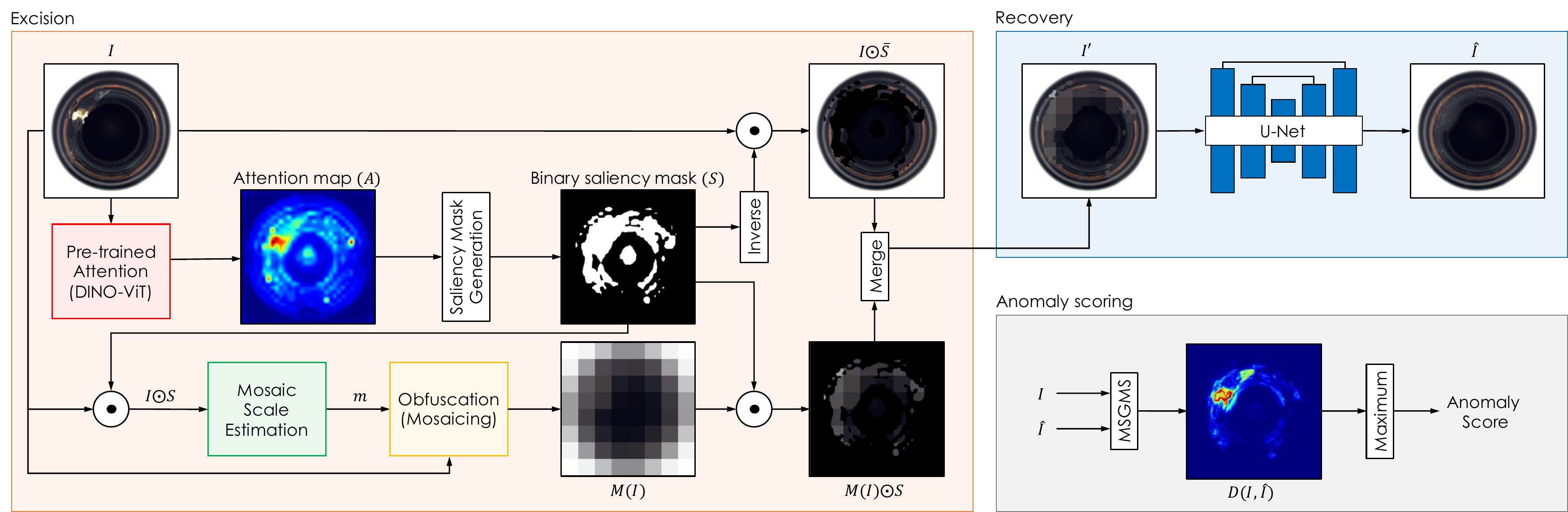}
    }
    \vspace{-0.6cm}
    \caption{An overview of EAR. EAR takes the reconstruction-by-inpainting approach and is characterized by single deterministic masking and visual obfuscation of masked regions for hint-providing. The orange box shows the excision process by exploiting the ImageNet~\cite{ImageNet_Deng_CVPR09} pre-trained DINO-ViT~\cite{DINO_Caron_ICCV21} to mask suspected defective regions. To promote the reconstruction of the region into a normal form, visually obfuscated information by mosaicing is provided as a hint. At this time, mosaic scale, $m$, is estimated from the saliency region of the given product image to provide a proper hint. Mosaic obfuscation is performed by average pooling with $m \times m$ pixels of image and upscaling it into the original scale. The blue box shows the recovery process that reconstructs the corrupted region in $I'$ into $\hat{I}$. Abnormality is decided based on a maximum value of $D(I, \hat{I})$ as shown in the gray box. Note that $D(I, \hat{I})$ is a distance map calculated by multi-scale gradient magnitude similarity (MSGMS)~\cite{RIAD_Vitzan_PR21}.}
    \label{fig:smore}
    \vspace{-0.5cm}
\end{figure*}

Recent years have witnessed a large amount of research efforts aimed at enhancing the UAD performance by exploring novel neural network (NN) structures and innovative training strategies. 
Those can be divided into two main categories: 
1) employing an additional module to existing NNs such as generative adversarial networks (GAN)~\cite{GANomaly_Akcay_ACCV18,fAno_Thomas_MIA19,HPGAN_Park_ETRIJ21,DGM_Tang_MIA21} or memory module~\cite{MemAE_Gong_ICCV19,BWMem_Hou_ICCV21,Mem_Kim_ICASSP23,SQUID_Xiang_CVPR23} and
2) changing of the training strategy to online knowledge distillation~\cite{RevDistill_Deng_CVPR22,SQUID_Xiang_CVPR23,MOKD_Song_CVPR23} or utilization of synthetic data~\cite{Draem_Vitzan_ICCV21,SGSF_Xing_arXiv22,DSR_Vitzan_ECCV22,MLDFR_Gui_TII23}.
Those methods successfully improve the performance by refining widely adopted mainstream NNs such as U-Net~\cite{Unet_Olaf_MICCAI15}.
However, amid the pursuit of ever-more sophisticated techniques to achieve better performance for specific benchmark datasets, their solutions have common limitation of increase of computational expense by employing large-scale deep NNs.

To avoid the above situations, the reconstruction-by-inpainting approach~\cite{RIAD_Vitzan_PR21,Randommask_Nardin_ICIAPW22,MaskSwin_Jiang_TII23,MAEMRI_Lang_arXiv23,SSM_Huang_TM22,Iter_Hitoshi_JIP22,InTra_Pirnay_ICIAP22,MSR_Cosmin_ICMLW23} have been investigated to improve the UAD performance without increasing the scale of the NN structure to use.
This approach fundamentally prevents accurate reconstruction of unseen anomalous patterns by making them not visible through masking.
However, there still remain the following problems to address:
1) inference latency due to multiple masking or progressive inpainting strategy~\cite{SupPix_Li_BMVC20,RIAD_Vitzan_PR21,SSM_Huang_TM22,Iter_Hitoshi_JIP22,InTra_Pirnay_ICIAP22,MSR_Cosmin_ICMLW23},
2) output inconsistency by random masking~\cite{RIAD_Vitzan_PR21,Randommask_Nardin_ICIAPW22,MaskSwin_Jiang_TII23,MAEMRI_Lang_arXiv23}, and
3) inaccurate reconstruction of normal patterns due to a large mask ratio~\cite{MAE_He_CVPR22,MAGE_Li_CVPR23}.

To solve these issues, we introduce a novel approach to enhance the UAD performance based on single deterministic masking. 
Our method, dubbed \textit{\textbf{E}xcision \textbf{A}nd \textbf{R}ecovery} (EAR), features attention-based visual defect obfuscation. 
That is, suspected defective regions are obfuscated by mosaicing as shown in Figure~\ref{fig:smore}.
EAR leverages the ImageNet~\cite{ImageNet_Deng_CVPR09} pre-trained DINO-ViT~\cite{DINO_Caron_ICCV21} that is known to have the ability to emphasize class-specific spatial features.
We exploit this property to highlight saliency regions within a given image and excise suspected anomalous regions for inpainting.
This deterministic single masking strategy allows fast processing and secures the output reliability.
Also, the problem of inaccurate reconstruction of the normal pattern due to large masked region is eased by the mosaic hint which is provided in the masked regions.
For this, we estimate the proper mosaic scale for the defective region by leveraging the ratio of principal curvatures of Hessian matrix, which was also used in scale-invariant feature transformation (SIFT)~\cite{SIFT_Lowe_IJCV04} to compute the degree of edge response.
Thereby, EAR achieves the UAD performance enhancement while it does not change the NN structure at all.
The details of the above design components will be described in Sections~\ref{subsec:salmask}, ~\ref{subsec:obfuscation}, and ~\ref{subsec:det_mos_scale} sequentially.

Experimental results with the public industrial visual inspection dataset, MVTec AD~\cite{MVTec_Bergmann_CVPR19}, demonstrate that EAR further enhances the UAD performance compared to the same or similar scale of NNs. 
Visual reconstruction results in Figures~\ref{fig:smore_ear} and \ref{fig:vis_comparison} indicate that EAR has desirable \textit{contained generalization ability}~\cite{LAMP_Park_arXiv23} for the UAD task.
That is, suspected defective regions that are visually obfuscated are reconstructed accurately when the input pattern is in the seen normal category, and that the reconstruction is inaccurate when the input includes unseen anomalous patterns.

Overall, our contributions are summarized as follows:
\vspace{-0.2cm}
\begin{itemize}
    \item The proposed pre-trained spatial attention-based single deterministic masking method has advanced the state-of-the-art methods in the reconstruction-by-inpainting approach for UAD, securing both higher throughput and output reliability.
    \vspace{-0.3cm}

    \item Our hint-providing strategy by visual obfuscation on masked regions further enhances the UAD performance with the proposed mosaic scale estimation method.
    \vspace{-0.3cm}
    
    \item Our method is distinguished from others by enhancing the UAD performance without any changes in the NN structure. This allows practical deployability of deep NNs in the industrial environments as an edge computing approach.
\end{itemize}

\section{Related works}

\subsection{Simple but powerful UAD models}
There have been efforts to enhance the UAD performance based on widely known NNs, such as auto-encoder (AE) or U-Net~\cite{Unet_Olaf_MICCAI15}, without changing much of their structure.
Among AE variants are, 
MS-CAM~\cite{MSCAM_Li_Sensors22} presents a multi-scale channel attention module with an adversarial learning strategy. 
GANomaly~\cite{GANomaly_Akcay_ACCV18} adopts feature distance loss to perform better normal pattern reconstruction.
SCADN~\cite{SCADN_Yan_AAAI21} performs multi-scale striped masking before feeding input to their NN.
In cases of U-Net~\cite{Unet_Olaf_MICCAI15} variants, DAAD~\cite{BWMem_Hou_ICCV21}  includes block-wise memory module, and RIAD~\cite{RIAD_Vitzan_PR21} proposes the reconstruction-by-inpainting strategy with multiple square patched disjoint masks.
These approaches maximize the UAD performance while keeping the scale of the NN relatively small. 

We also employ an U-Net~\cite{Unet_Olaf_MICCAI15} structure and at the same time, pursue a practically deployable solution that allows NNs to operate properly in industrial environments as a way for edge computing.

\subsection{Reconstruction-by-inpainting methods}
UAD based on reconstruction-by-inpainting is an effective self-supervision technique for representation learning to prevent an UAD model from accurately reconstructing unseen anomalous patterns~\cite{RIAD_Vitzan_PR21,Randommask_Nardin_ICIAPW22,MaskSwin_Jiang_TII23,MAEMRI_Lang_arXiv23,SSM_Huang_TM22,Iter_Hitoshi_JIP22,InTra_Pirnay_ICIAP22,MSR_Cosmin_ICMLW23}.
Specifically, methods such as random masking~\cite{RIAD_Vitzan_PR21,Randommask_Nardin_ICIAPW22,MaskSwin_Jiang_TII23,MAEMRI_Lang_arXiv23}, multiple disjoint masking~\cite{RIAD_Vitzan_PR21,SSM_Huang_TM22}, and progressive inpainting from the initial masks~\cite{SupPix_Li_BMVC20,Iter_Hitoshi_JIP22,SSM_Huang_TM22,InTra_Pirnay_ICIAP22,MSR_Cosmin_ICMLW23} have been developed.

The common limitation of multiple masking and progressive inpainting is inference latency due to the multiple inferences.
In addition, the random masking strategy causes the problem of output inconsistency when applied to the reconstruction-by-inpainting approach.
Thus, to develop a practically deployable solution for ensuring real-time defect detection and output reliability, we should consider the following:
1) deterministic mask generation strategy,
2) minimizing the number of masks, and
3) immediate inpainting strategy rather than a progressive inpainting strategy.

To meet the above requirements, we exploit a pre-trained attention model for deterministic single masking.
The deterministic single masking strategy allows real-time processing and, at the same time, secures the output reliability.

\subsection{Hint-providing strategies for masked regions}
Researches report that attention-based saliency masking~\cite{AttMask_Kakogeorgiou_ECCV22,AMT_Liu_AAAI23} or non-saliency masking~\cite{APMask_Bozorgtabar_AAAI23,OoD_Sim_BigComp23} is more effective and helpful for representation learning. 
Their intention is to eliminate unnecessary input information for their objective, representation learning or object recognition.

However, since those masking methods will empty all the information in the suspected anomalous regions, accurate reconstruction of normal patterns becomes hard, especially when the masked region is large.
To ease this situation, an additional strategy that randomly leaves a few patches within masked saliency areas as hint information for reconstruction can be considered~\cite{AttMask_Kakogeorgiou_ECCV22,AMT_Liu_AAAI23}.
This strategy serves to provide initial information for inpainting the masked regions and accurate reconstruction.
However, the randomness of their patch hint-providing causes the output inconsistency problem.

We present a visual obfuscation-based hint-providing scheme to promote the accurate reconstruction of normal patterns.

\vspace{-0.15cm}
\section{Methods}

\subsection{Overview}
Due to the class imbalance problem stemming from the scarcity of abnormal situations, we adopt a self-supervised learning strategy to conduct target representation learning of normal samples.
An overall schematic diagram of EAR is shown in Figure~\ref{fig:smore}. 
The excision stage is composed of two steps.
First, we generate a deterministic single saliency mask, $S$, from attention map, $A$, by exploiting the ImageNet~\cite{ImageNet_Deng_CVPR09} pre-trained DINO-ViT~\cite{DINO_Caron_ICCV21}.
The resulting saliency mask, $S$, indicates suspected anomalous regions.
Then, we provide a mosaic hint of the masked regions for reconstruction.
To provide a proper hint by obfuscation, mosaic scale, $m$, is estimated from the part of the given product image that correspond to the saliency region.
The result of recovery, $\hat{I}$, will be obtained by feeding $I'$ into the U-Net~\cite{Unet_Olaf_MICCAI15}.
The magnitude of the reconstruction error, especially the maximum value of multi-scale gradient magnitude similarity (MSGMS)~\cite{RIAD_Vitzan_PR21}, between $I$ and $\hat{I}$ is used to determine whether the product is defective or not.

\subsection{Saliency mask generation}
\label{subsec:salmask}
We aim to develop a real-time and reliable solution by avoiding inference latency and output inconsistency.
For this, we propose a deterministic saliency masking strategy by exploiting a pre-trained self-attention model. 
Specifically, the ImageNet~\cite{ImageNet_Deng_CVPR09} pre-trained DINO-ViT~\cite{DINO_Caron_ICCV21} is used in this study which is trained with a self-distillation strategy.
First, we feed an input image $I$ into the DINO-ViT~\cite{DINO_Caron_ICCV21} and get an attention map, $A$, by averaging [CLS] tokens, multi-heads of the last layer.
Then, we binarize the attention map, $A$, by thresholding the upper quartile value $\mu+0.674\sigma$ ($Q3$)~\cite{Fence_Tukey_Book77} of pixel-wise attention scores to generate a binary saliency mask, $S$.
Referring to the probable error~\cite{ProbableError_Dodge_Oxford03}, we regard upper $Q3$ values as suspected anomalous regions.
This allows the mask size to be sufficiently large enough to cover suspected anomalous regions while keeping it reasonably small.

$S$ is used to cut out the suspected anomalous regions in normal samples in training, and the UAD model will be optimized to inpaint the empty region.
After training, the recovered masked region by the UAD model will be accurately matched when the $I$ does not include any unseen anomalous patterns.
However, if the masked region covers anomalous patterns, the UAD model will struggle to recover its original defective form.
Therefore, defective products can be effectively detected due to relatively large reconstruction errors.

\subsection{Obfuscation-based hint for reconstruction}
\label{subsec:obfuscation}
Saliency masking empties the defective information in the suspected defective regions to help transform masked unseen anomalous regions into normal forms.
However, not leaving any clues in the masked region could cause inaccurate reconstruction of normal patterns, degrading the UAD performance.

We propose a hint-providing strategy with visual obfuscation on masked saliency regions for accurate reconstruction of normal patterns.
For visual obfuscation, we adopt mosaicing of proper scale depending on the defect scale.
For mosaicing, we create each single representative value within each square patch of $m \times m$ pixels by average pooling.
Thus, the mosaic scale is represented by $m$.
Determining a proper mosaic scale is described in Section~\ref{subsec:det_mos_scale}.
The average pooled image is upscaled into the original scale with the nearest interpolation and combined with a saliency mask to provide the masked regions with the proper mosaic hint as shown in Figure~\ref{fig:smore}.
When the mosaic method described above is denoted by $M$, the hint-providing method is expressed as \eqref{eq:processing}.
The processed image $I'$ will be fed into the UAD model for reconstruction.

\begin{equation}
    \begin{aligned}
        I' = M(I)\odot{}S + I\odot{}\bar{S}
    \end{aligned}
    \label{eq:processing}
\end{equation}

This mosaicing with the proper mosaic scale makes anomalous regions visually obfuscated to an extent that helps efficient reconstruction of normal patterns, and contains accurate reconstruction of anomalous patterns.

\subsection{Determining mosaic scale}
\label{subsec:det_mos_scale}
Depending on the mosaic scale, there is a difference in the information details of provided hints for masked regions.
Since the mosaic scale to use is a factor that determines the reconstruction quality, it directly affects the UAD performance.
As the optimal scale of the mosaic for each product is not known in advance, estimating the proper mosaic scale is necessary to give the best possible hint.

To construct a mosaic scale estimation model, we first obtain the optimal mosaic scale $m^{*}$ for each product in the MVTec AD dataset~\cite{MVTec_Bergmann_CVPR19} through grid search.
For setting a representative value for each product, we exploit the degree of edge response, $r$, in \eqref{eq:r_eigen}, which was used in SIFT~\cite{SIFT_Lowe_IJCV04}.

\begin{equation}
    \begin{aligned}
        r = Tr(H)^{2} / Det(H), \quad{}
        H = \nabla{}^{2}I
    \end{aligned}
    \label{eq:r_eigen}
\end{equation}

\begin{figure}[t]
    \resizebox{\columnwidth}{!}{%
        \begin{tabular}{cc}
            \includegraphics*[width=0.7\linewidth,trim={0.3cm 0.0cm 0.3cm 0.0cm},clip]{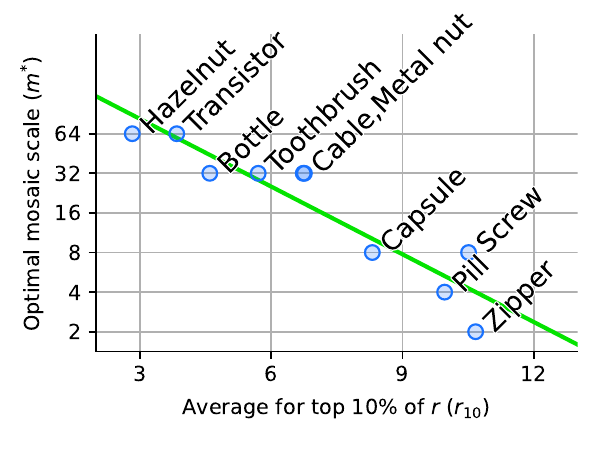} & 
            \includegraphics*[width=0.7\linewidth,trim={0.3cm 0.0cm 0.3cm 0.0cm},clip]{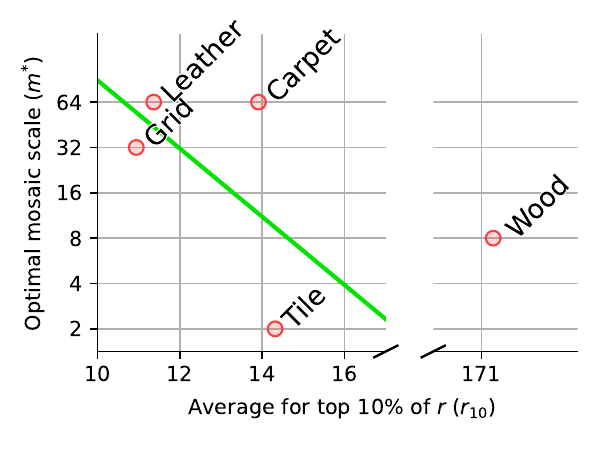} \\ 
        \end{tabular}
    }
    \vspace{-0.4cm}
    \caption{Linear regression model between $r_{10}$ and $m^{*}$. $m^{*}$ found by grid search is denoted by blue and red circles for $r_{10}$, and their correlation coefficients are -0.939 and -0.497 for 10 object subsets and 5 texture subsets, respectively. The linear function $f$ to estimate $\hat{m}$ is shown by the green line. We determine $\hat{m}$ by quantizing $f(r_{10})$ to the nearest power of 2. As an exception, $\hat{m}$ for `wood' is set to 2, which is the smallest mosaic scale, due to its original estimation being a negative value.}
    \label{fig:estimation}
    \vspace{-0.3cm}
\end{figure}

\begin{figure}[t]
    \setlength{\tabcolsep}{0pt}
    \resizebox{\linewidth}{!}{%
        \begin{tabular}{ccc}
            \includegraphics*[width=0.35\columnwidth,trim={0.0cm 0.4cm 0.3cm 0.4cm},clip]{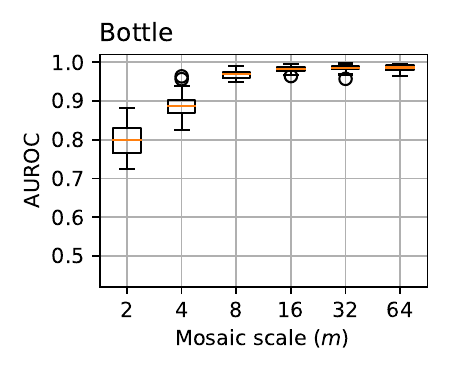} &
            \includegraphics*[width=0.35\columnwidth,trim={0.0cm 0.4cm 0.3cm 0.4cm},clip]{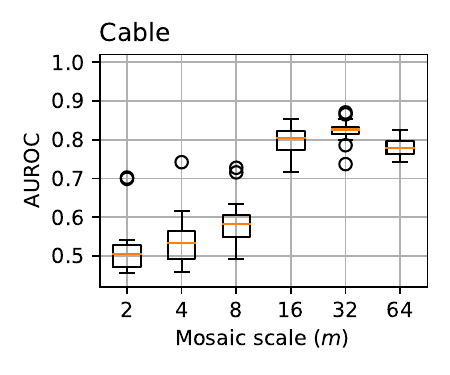} &
            \includegraphics*[width=0.35\columnwidth,trim={0.0cm 0.4cm 0.3cm 0.4cm},clip]{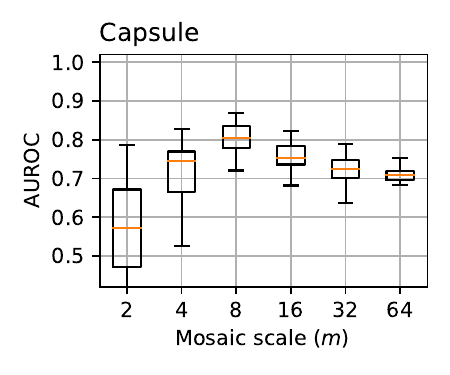} \\
            
            \includegraphics*[width=0.35\columnwidth,trim={0.0cm 0.4cm 0.3cm 0.4cm},clip]{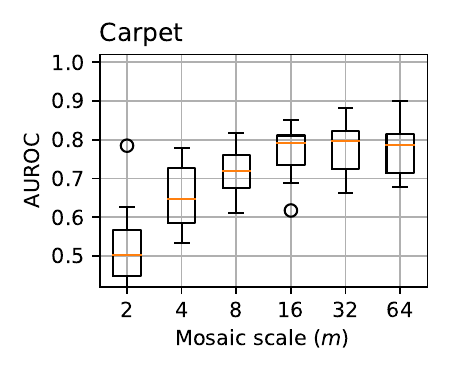} &
            \includegraphics*[width=0.35\columnwidth,trim={0.0cm 0.4cm 0.3cm 0.4cm},clip]{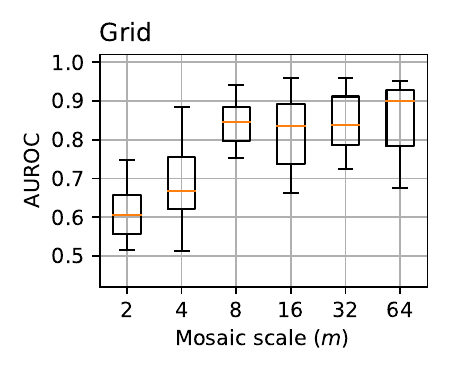} &
            \includegraphics*[width=0.35\columnwidth,trim={0.0cm 0.4cm 0.3cm 0.4cm},clip]{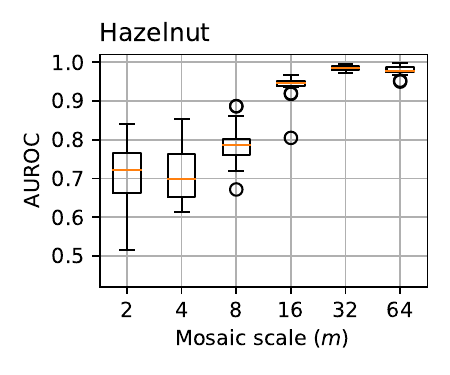} \\
            
            \includegraphics*[width=0.35\columnwidth,trim={0.0cm 0.4cm 0.3cm 0.4cm},clip]{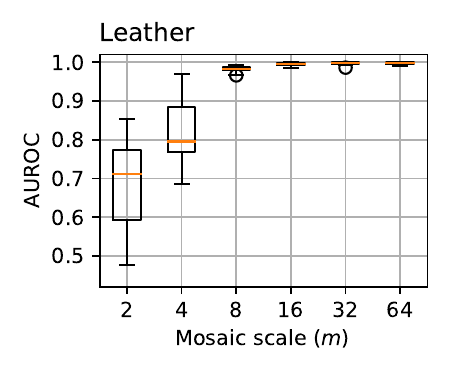} &
            \includegraphics*[width=0.35\columnwidth,trim={0.0cm 0.4cm 0.3cm 0.4cm},clip]{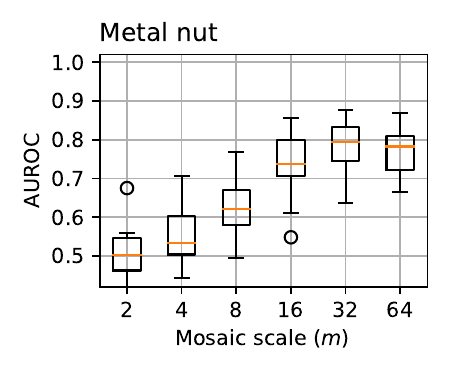} &
            \includegraphics*[width=0.35\columnwidth,trim={0.0cm 0.4cm 0.3cm 0.4cm},clip]{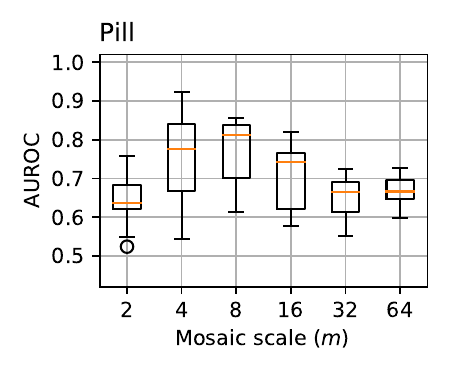} \\
            
            \includegraphics*[width=0.35\columnwidth,trim={0.0cm 0.4cm 0.3cm 0.4cm},clip]{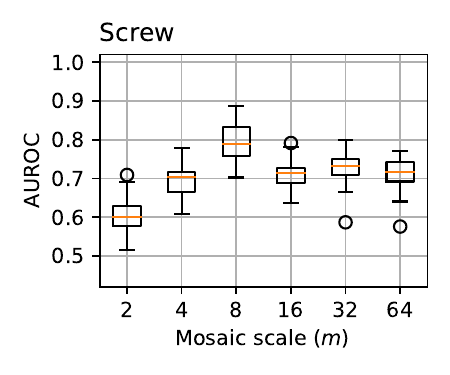} &
            \includegraphics*[width=0.35\columnwidth,trim={0.0cm 0.4cm 0.3cm 0.4cm},clip]{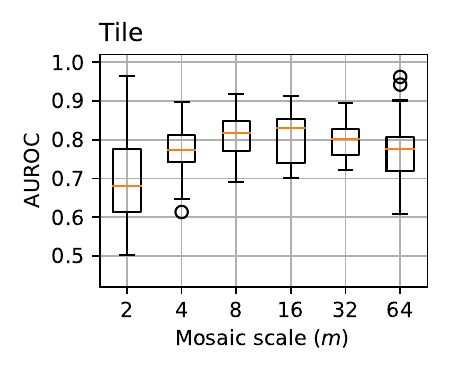} &
            \includegraphics*[width=0.35\columnwidth,trim={0.0cm 0.4cm 0.3cm 0.4cm},clip]{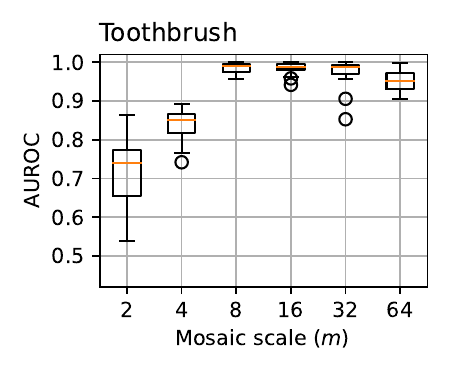} \\
            
            \includegraphics*[width=0.35\columnwidth,trim={0.0cm 0.4cm 0.3cm 0.4cm},clip]{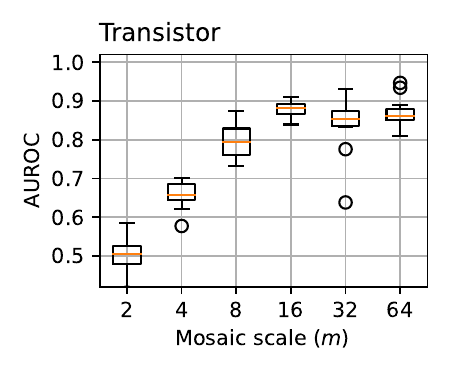} &
            \includegraphics*[width=0.35\columnwidth,trim={0.0cm 0.4cm 0.3cm 0.4cm},clip]{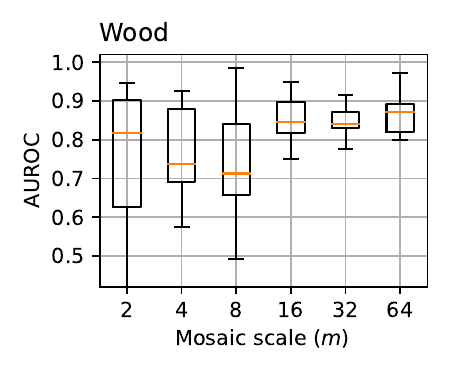} &
            \includegraphics*[width=0.35\columnwidth,trim={0.0cm 0.4cm 0.3cm 0.4cm},clip]{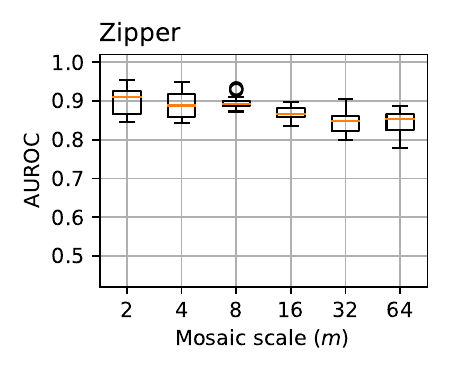} \\
        \end{tabular}
    }
    \vspace{-0.3cm}
    \caption{Grid search results of finding the optimal mosaic scale $m^{*}$ for hint-providing. Overall, it appears that a larger mosaic scale is advantageous in obtaining a higher AUROC. However, for products with detailed patterns such as capsule, pill, and screw, a moderately small mosaic scale is recommended. The correlation between the visual characteristic of the product and the optimal mosaic scale is shown in Figure~\ref{fig:estimation}.}
    \label{fig:opt_mosaic}
    \vspace{-0.4cm}
\end{figure}

\begin{table}[t]
    \centering
    \scriptsize
    \caption{Summary of optimal mosaic scale $m^{*}$ and estimated mosaic scale $\hat{m}$. $m^{*}$ is found by grid search, and $\hat{m}$ is estimated mosaic scale determined by quantizing $f(r_{10})$ to the nearest power of 2. The linear regression function $f$ for each object and texture is shown in Figure~\ref{fig:estimation}.}
    \setlength\tabcolsep{3pt}
        \begin{tabular}{l|cc c l|cc}
            \cline{1-3}
            \cline{5-7}
                \multicolumn{3}{c}{Objects} & \quad{} & \multicolumn{3}{c}{Textures} \\
            \cline{1-3}
            \cline{5-7}
                Product & $m^{*}$ & $\hat{m}$ & \quad{} & Product & $m^{*}$ & $\hat{m}$ \\
            \hhline{===~===}
                Bottle & 32 & 32 & \quad{} & Carpet & 64 & 16 \\
                Cable & 32 & 16 & \quad{} & Grid & 32 & 64 \\
                Capsule & 8 & 8 & \quad{} & Leather & 64 & 32 \\
                Hazelnut & 64 & 64 & \quad{} & Tile & 2 & 8 \\
                Metal nut & 32 & 16 & \quad{} & Wood & 8 & 2 \\
            \cline{5-7}
                Pill & 4 & 4 \\
                Screw & 8 & 4 \\
                Toothbrush & 32 & 32 \\
                Transistor & 64 & 64 \\
                Zipper & 2 & 4 \\
            \cline{1-3}
        \end{tabular}
    \label{table:opt_mosaic}
    \vspace{-0.3cm}
\end{table}

\begin{table}[t]
    \centering
    \footnotesize
    \caption{Summary of the tuned hyperparameters and their values. We explore the optimal combination of hyperparameters in a grid search manner.}
    \setlength\tabcolsep{2pt}
        \begin{tabular}{l||l}
        \hline
            Hyperparameter & Values \\
        \hline
        \hline
            Kernel size ($k$) & 3 and 5\\
            Learning rate ($\eta$) & 1e-3, 1e-4, and 1e-5\\
            Learning rate scheduling & fixed, warm-up~\cite{Warm_Goyal_arXiv17} and SGDR~\cite{SGDR_Ilya_ICLR17}\\
        \hline
        \end{tabular}
    \label{table:hyperparameter}
    \vspace{-0.3cm}
\end{table}

For summarizing the pixel-wise edge response, we only take the saliency region in $I$ and average the top 10\% of them.
Let us denote this by $r_{10}$.
We could successfully relate $r_{10}$ to $m^{*}$ and their linear relation is shown in Figure~\ref{fig:estimation}.
They show a strong correlation.
Products with detailed features or rough surfaces give high values of $r_{10}$ while products with relatively smooth surfaces show low values.

We optimize the linear function of the mosaic scale estimation model for each object and texture subset.
An estimated mosaic scale, $\hat{m}$, is determined by quantizing $f(r_{10})$ to the nearest power of 2.
For experiments, $\hat{m}$ will be used for EAR training, and the results using $m^{*}$ will also be presented to verify the effectiveness of the proposed mosaic scale estimation method.

\subsection{Training objectives}
A prior study~\cite{RIAD_Vitzan_PR21} has shown a satisfactory result to detect various-sized defects by employing MSGMS as in~\eqref{eq:loss_msgms}. 
Their training objective also includes $\mathcal{L}_{2}$ (pixel-wise distance) and structural similarity index measure (SSIM)~\cite{SSIM_Wang_TIP2004} which are widely used for training of a reconstruction model.
We also inherit the above for training and anomaly scoring.
In MSGMS, we set multiple scales $N$ to 3.
\vspace{-0.3cm}



\begin{equation}
    \begin{aligned}
        \mathcal{L}_{msgms}(I, \hat{I}) = \sum_{n=1}^{N} \left( 1 - \frac{2g(I^{n})g(\hat{I}^{n}) + c}{g(I^{n})^2 + g(\hat{I}^{n})^2 + c} \right)
    \end{aligned}
    \label{eq:loss_msgms}
\end{equation}

\begin{equation}
    \begin{aligned}
        \mathcal{L}_{comb} = \lambda_{2}\mathcal{L}_{2} + \lambda_{ssim}\mathcal{L}_{ssim} + \lambda_{msgms}\mathcal{L}_{msgms}
    \end{aligned}
    \label{eq:loss_combine}
\end{equation}

Three loss terms $\mathcal{L}_{2}$, $\mathcal{L}_{ssim}$, and $\mathcal{L}_{msgms}$ are combined by applying weights $\lambda$ as~\eqref{eq:loss_combine}.
Then, we apply a loss transformation method LAMP~\cite{LAMP_Park_arXiv23} on~\eqref{eq:loss_combine}. 
LAMP~\cite{LAMP_Park_arXiv23} is known to enhance the UAD performance by only loss amplification of the training process.
In addition, it can be applied to any UAD training process because it does not depend on NN structures or preprocessing methods.
The final loss function for training EAR is \eqref{eq:loss_training}.

\begin{equation}
    \begin{aligned}
        \mathcal{L}_{comb}^{LAMP}(I, \hat{I}) = -\log\Bigl(1 - \mathcal{L}_{comb}(I, \hat{I})\Bigr)
    \end{aligned}
    \label{eq:loss_training}
\end{equation}

\section{Experiments}

\subsection{Experimental setup}
To evaluate the performance of EAR, we use the public industrial visual inspection dataset, MVTec AD~\cite{MVTec_Bergmann_CVPR19}, for the experiments.
MVTec AD~\cite{MVTec_Bergmann_CVPR19} provides a total of 15 subtasks with 10 objects and 5 textures.
Each training set for these tasks only provides normal, anomaly-free samples.
The test set includes both normal and defective samples.

\begin{figure}[t]
    \resizebox{\columnwidth}{!}{%
        \includegraphics*[width=1.0\linewidth,trim={0.0cm 0.0cm 0.0cm 0.0cm},clip]{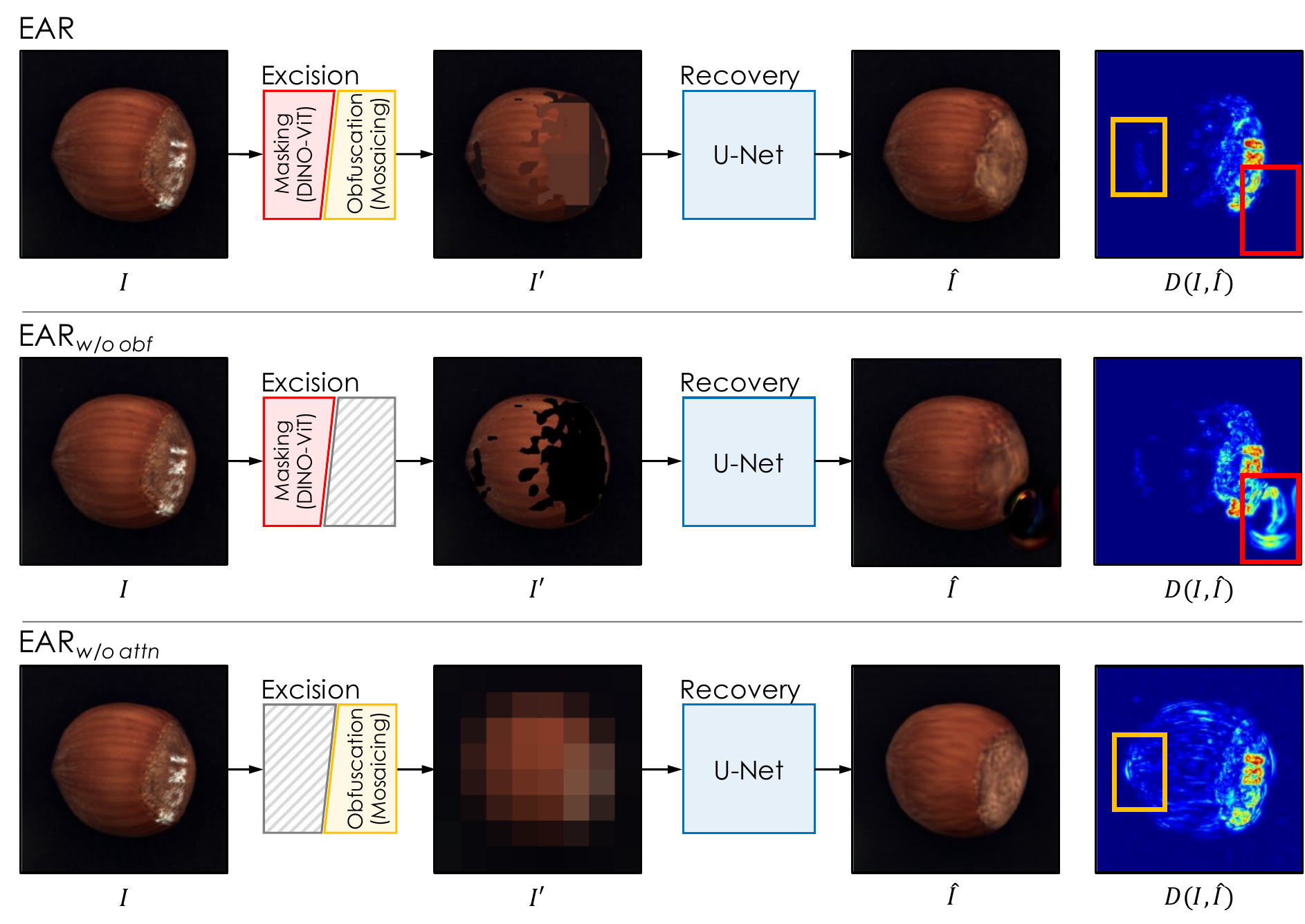}
    }
    \vspace{-0.6cm}
    \caption{Visual comparison of the results when disabling each design component of EAR: visual obfuscation by mosaicing and saliency masking. The EAR variants to confirm the aforementioned effect are the following: 1) EAR is a full model that activates all the components. 2) EAR$_{\mbox{\textit{w/o obf}}}$ does not provide visual obfuscation-based hints on masked regions. 3) EAR$_{\mbox{\textit{w/o attn}}}$ disables the masking component which exploits the ImageNet~\cite{ImageNet_Deng_CVPR09} pre-trained DINO-ViT~\cite{DINO_Caron_ICCV21}. A full model of EAR accurately reconstructs normal regions within a defective sample, marked in the yellow box. In contrast, anomalous regions, marked in the red box, are transformed into a normal form and yield a large reconstruction error. EAR$_{\mbox{\textit{w/o obf}}}$ and EAR$_{\mbox{\textit{w/o attn}}}$ cases show inaccurate inpainting compared to EAR. Best viewed in color.}
    \label{fig:smore_ear}
    \vspace{-0.2cm}
\end{figure}

\begin{figure*}[t]
    \scriptsize
    \setlength{\tabcolsep}{0pt}
    \resizebox{\linewidth}{!}{%
        \begin{tabular}{ccc cccc cccc cccc cccc}
                & & Input & & \multicolumn{3}{c}{RIAD~\cite{RIAD_Vitzan_PR21}} & & \multicolumn{3}{c}{EAR$_{\mbox{\textit{w/o obf}}}$} & & \multicolumn{3}{c}{EAR$_{\mbox{\textit{w/o attn}}}$} & & \multicolumn{3}{c}{EAR} \\
                \vspace{-0.20cm} \\
                & \:\: & $I$ & \:\: & $I'$ & $\hat{I}$ & $D(I,\hat{I})$ & \:\: & $I'$ & $\hat{I}$ & $D(I,\hat{I})$ & \:\: & $I'$ & $\hat{I}$ & $D(I,\hat{I})$ & \:\: & $I'$ & $\hat{I}$ & $D(I,\hat{I})$ \\

                \rotatebox[origin=c]{90}{\qquad{}\qquad{}\qquad{}\quad{}Normal} & & 
                \includegraphics*[width=0.21\columnwidth]{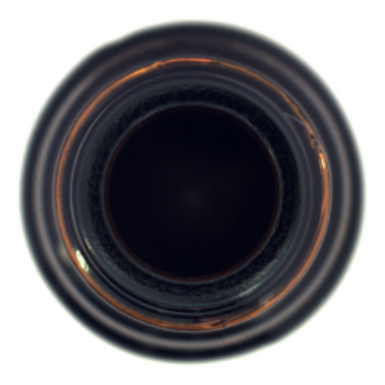} & &
                
                \includegraphics*[width=0.21\columnwidth]{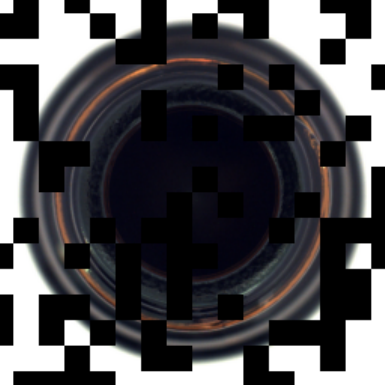} &
                \includegraphics*[width=0.21\columnwidth]{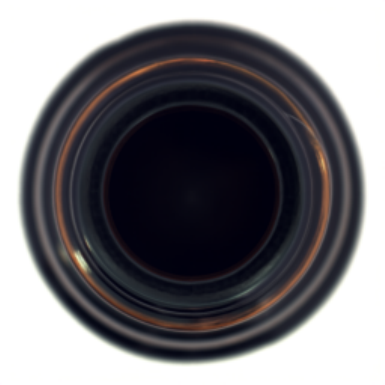} & 
                \includegraphics*[width=0.21\columnwidth]{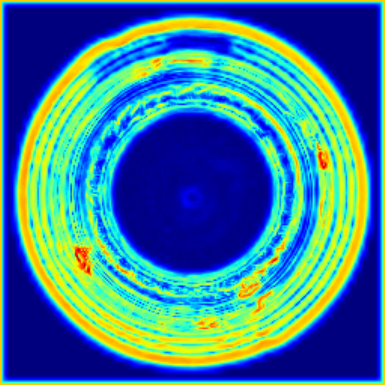} & &
    
                \includegraphics*[width=0.21\columnwidth]{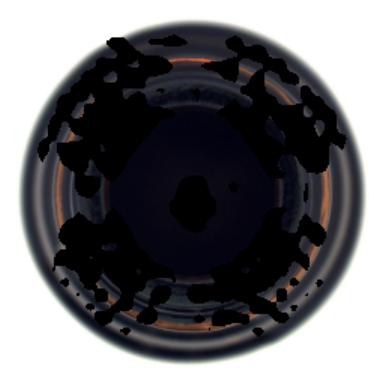} &
                \includegraphics*[width=0.21\columnwidth]{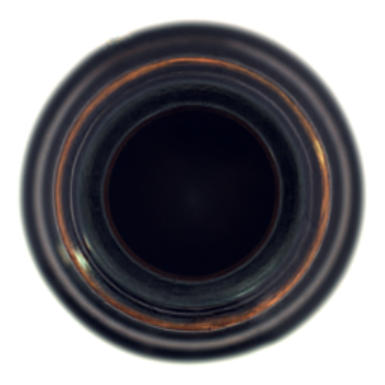} &
                \includegraphics*[width=0.21\columnwidth]{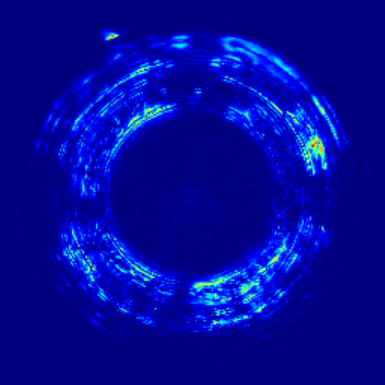} & &

                \includegraphics*[width=0.21\columnwidth]{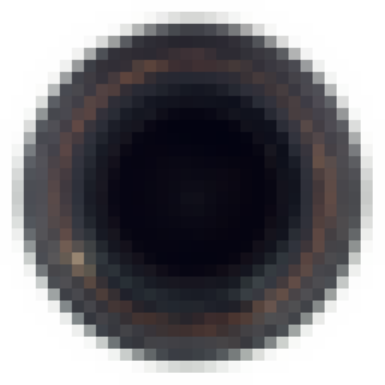} &
                \includegraphics*[width=0.21\columnwidth]{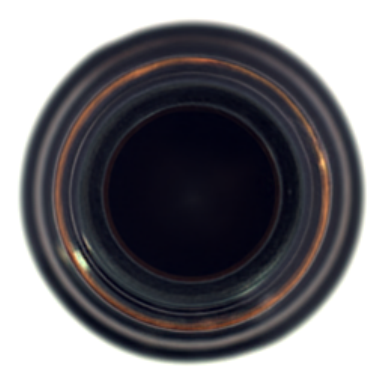} &
                \includegraphics*[width=0.21\columnwidth]{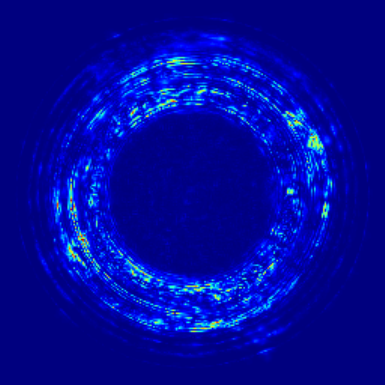} & &

                \includegraphics*[width=0.21\columnwidth]{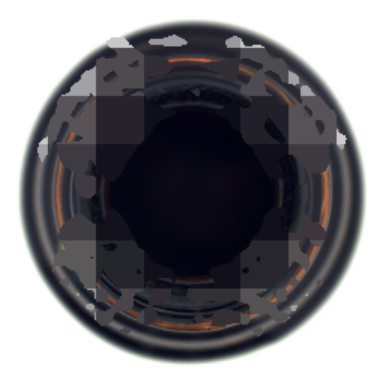} &
                \includegraphics*[width=0.21\columnwidth]{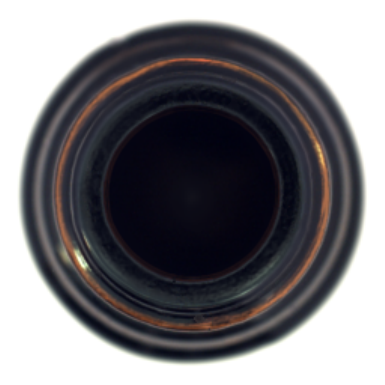} &
                \includegraphics*[width=0.21\columnwidth]{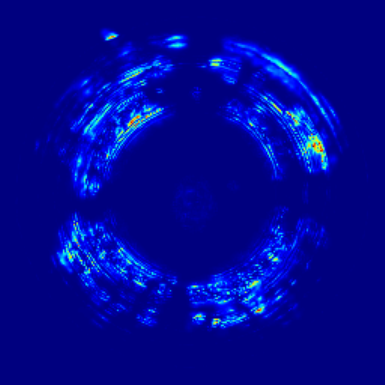} \\
                \vspace{-1.5cm} \\
    
                \rotatebox[origin=c]{90}{\qquad{}\qquad{}\qquad{}\quad{}Defective} & &
                \includegraphics*[width=0.21\columnwidth]{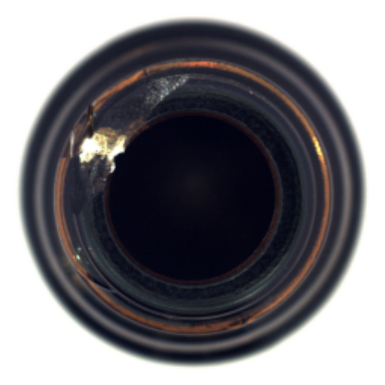} & &
                
                \includegraphics*[width=0.21\columnwidth]{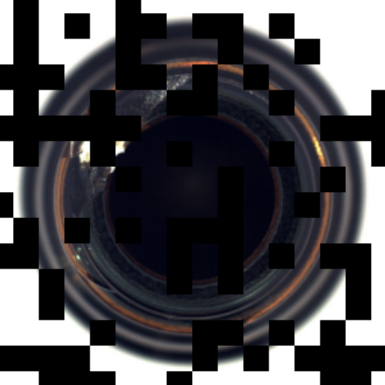} &
                \includegraphics*[width=0.21\columnwidth]{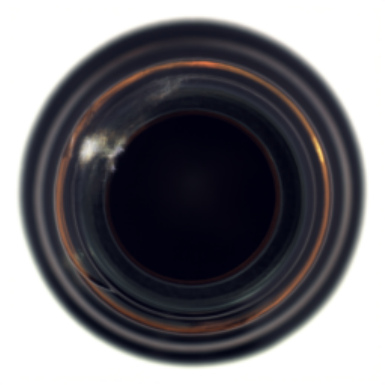} & 
                \includegraphics*[width=0.21\columnwidth]{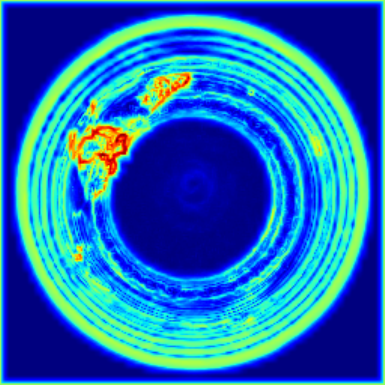} & &
    
                \includegraphics*[width=0.21\columnwidth]{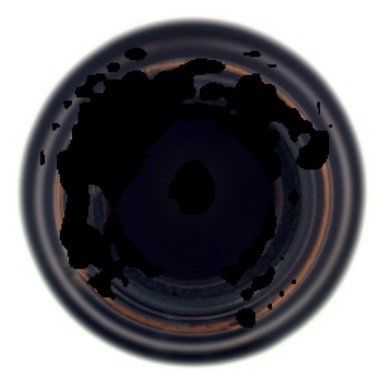} &
                \includegraphics*[width=0.21\columnwidth]{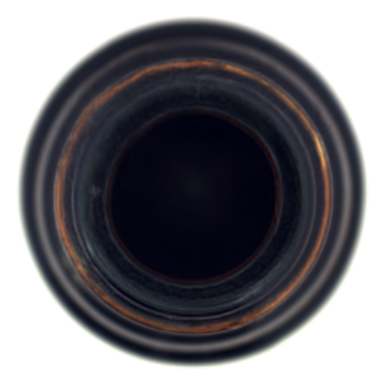} & 
                \includegraphics*[width=0.21\columnwidth]{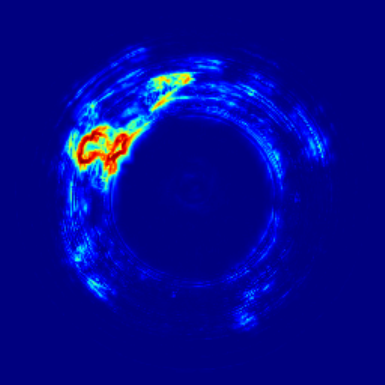} & &

                \includegraphics*[width=0.21\columnwidth]{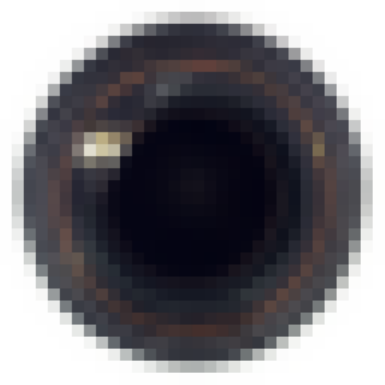} &
                \includegraphics*[width=0.21\columnwidth]{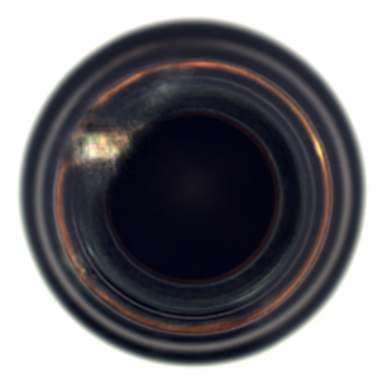} & 
                \includegraphics*[width=0.21\columnwidth]{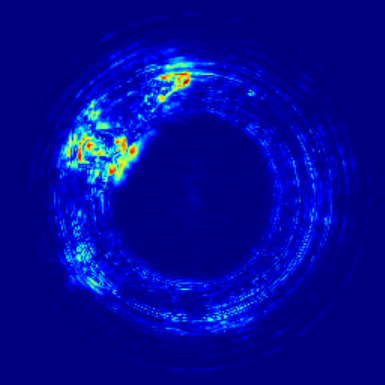} & &
    
                \includegraphics*[width=0.21\columnwidth]{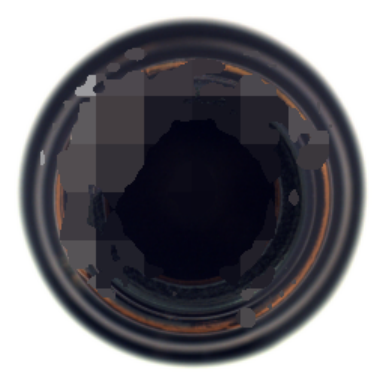} &
                \includegraphics*[width=0.21\columnwidth]{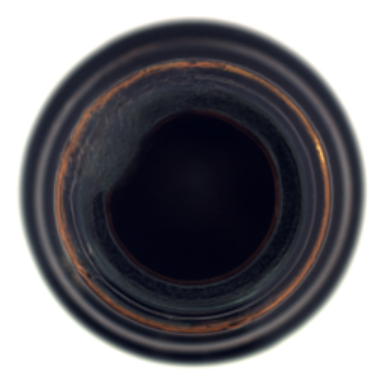} &
                \includegraphics*[width=0.21\columnwidth]{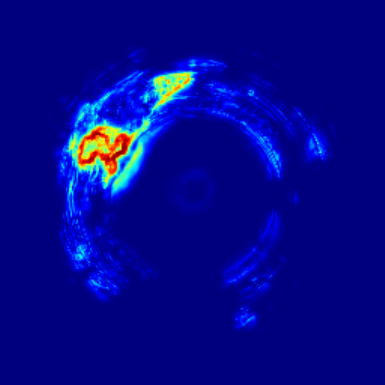} \\
                \vspace{-1.5cm} \\

            \hline
                \rotatebox[origin=c]{90}{\qquad{}\qquad{}\qquad{}\quad{}Normal} & & 
                \includegraphics*[width=0.21\columnwidth]{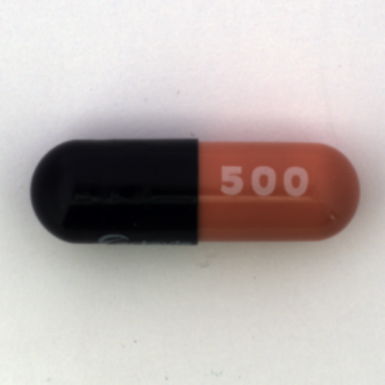} & &
                
                \includegraphics*[width=0.21\columnwidth]{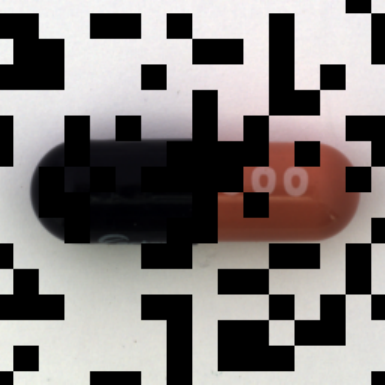} &
                \includegraphics*[width=0.21\columnwidth]{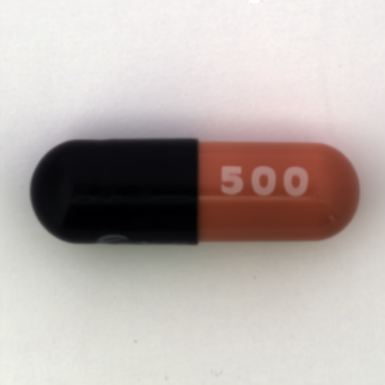} & 
                \includegraphics*[width=0.21\columnwidth]{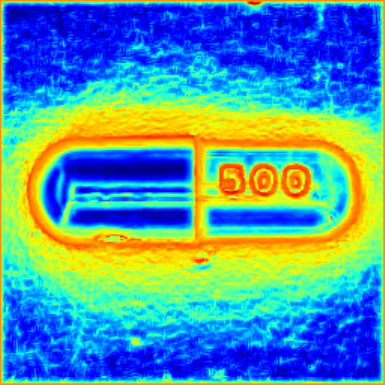} & &
    
                \includegraphics*[width=0.21\columnwidth]{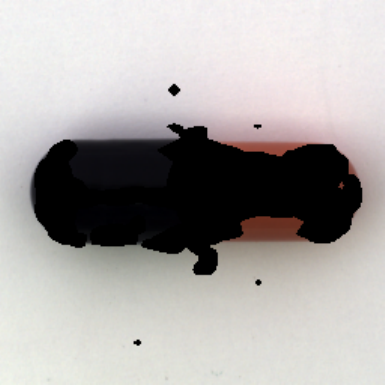} &
                \includegraphics*[width=0.21\columnwidth]{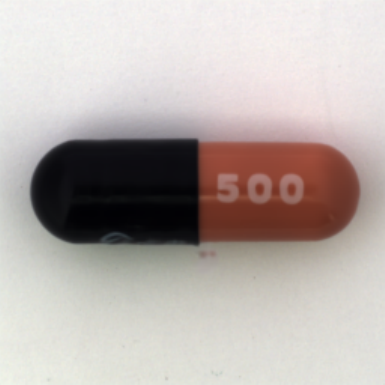} &
                \includegraphics*[width=0.21\columnwidth]{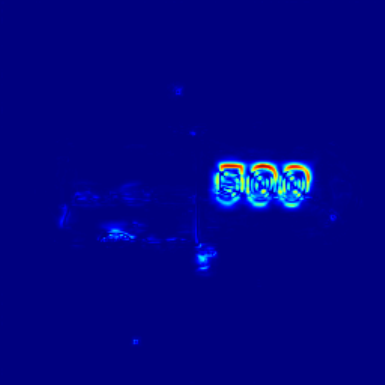} & &

                \includegraphics*[width=0.21\columnwidth]{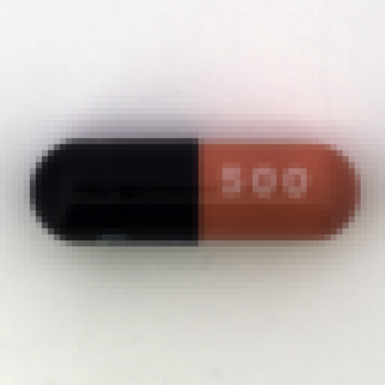} &
                \includegraphics*[width=0.21\columnwidth]{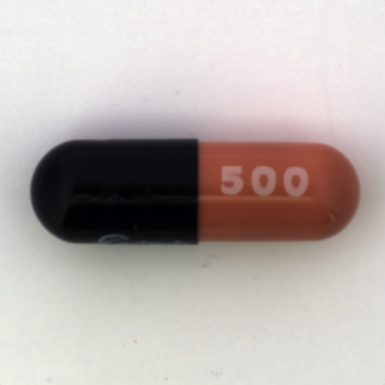} &
                \includegraphics*[width=0.21\columnwidth]{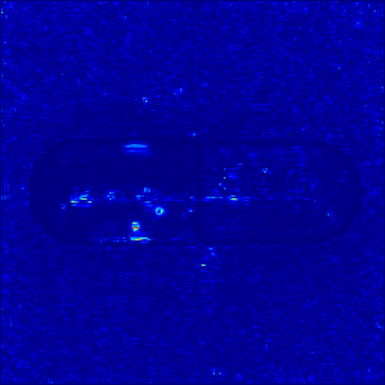} & &

                \includegraphics*[width=0.21\columnwidth]{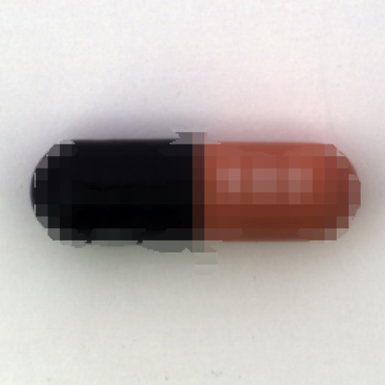} &
                \includegraphics*[width=0.21\columnwidth]{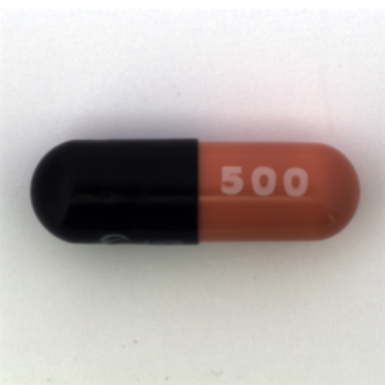} &
                \includegraphics*[width=0.21\columnwidth]{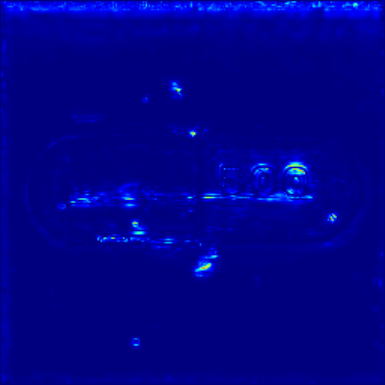} \\
                \vspace{-1.5cm} \\
    
                \rotatebox[origin=c]{90}{\qquad{}\qquad{}\qquad{}\quad{}Defective} & &
                \includegraphics*[width=0.21\columnwidth]{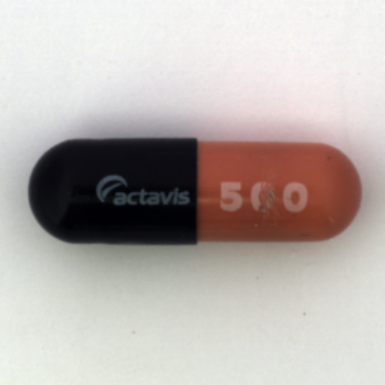} & &
                
                \includegraphics*[width=0.21\columnwidth]{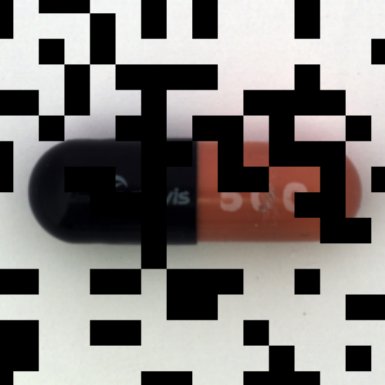} &
                \includegraphics*[width=0.21\columnwidth]{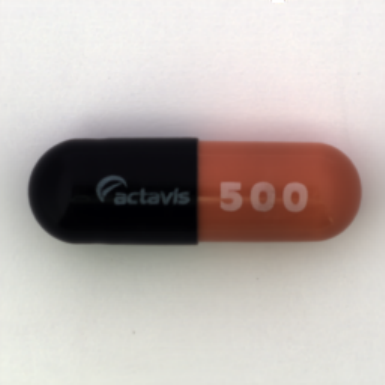} & 
                \includegraphics*[width=0.21\columnwidth]{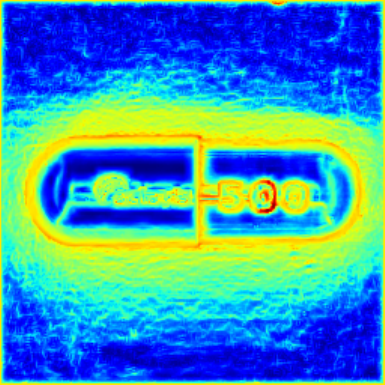} & &
    
                \includegraphics*[width=0.21\columnwidth]{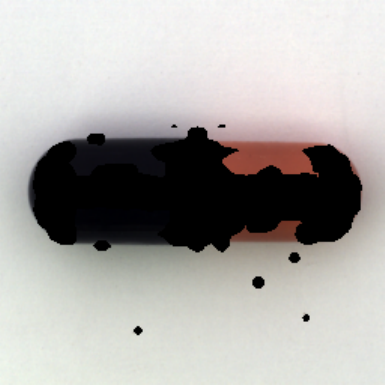} &
                \includegraphics*[width=0.21\columnwidth]{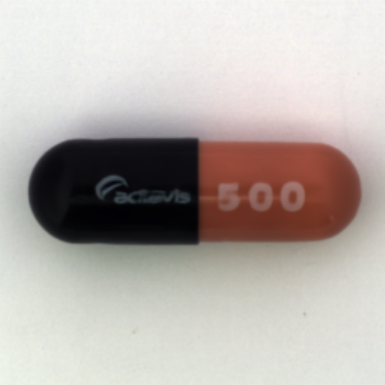} & 
                \includegraphics*[width=0.21\columnwidth]{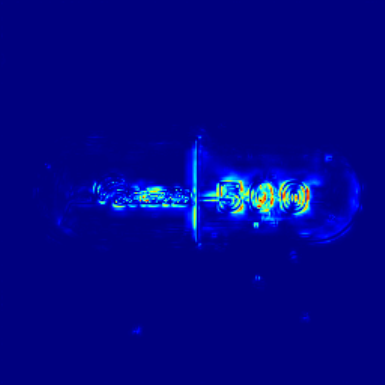} & &

                \includegraphics*[width=0.21\columnwidth]{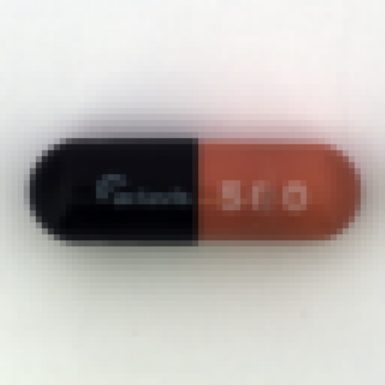} &
                \includegraphics*[width=0.21\columnwidth]{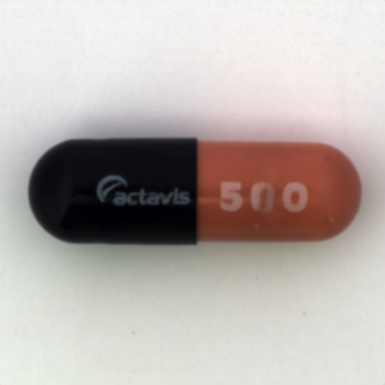} & 
                \includegraphics*[width=0.21\columnwidth]{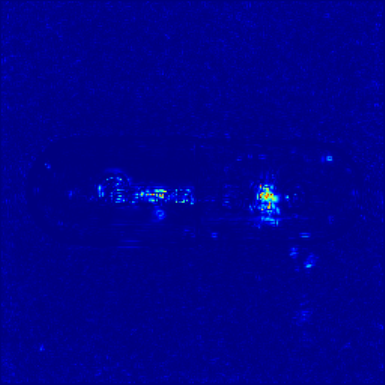} & &
    
                \includegraphics*[width=0.21\columnwidth]{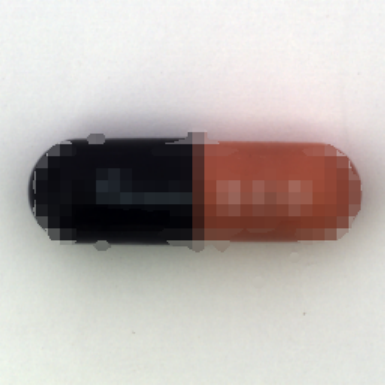} &
                \includegraphics*[width=0.21\columnwidth]{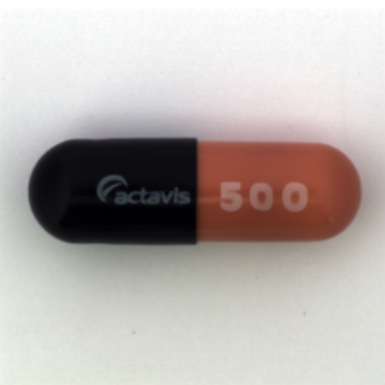} &
                \includegraphics*[width=0.21\columnwidth]{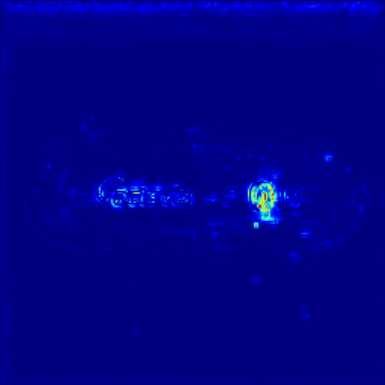} \\
                \vspace{-1.5cm} \\

            \hline
                \vspace{-0.25cm} \\
                \rotatebox[origin=c]{90}{\qquad{}\qquad{}\qquad{}\quad{}Normal} & &
                \includegraphics*[width=0.21\columnwidth]{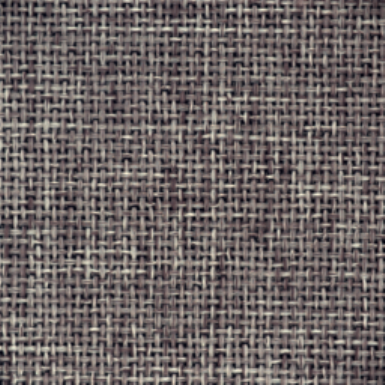} & &
                
                \includegraphics*[width=0.21\columnwidth]{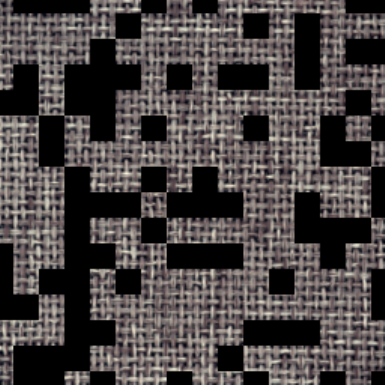} &
                \includegraphics*[width=0.21\columnwidth]{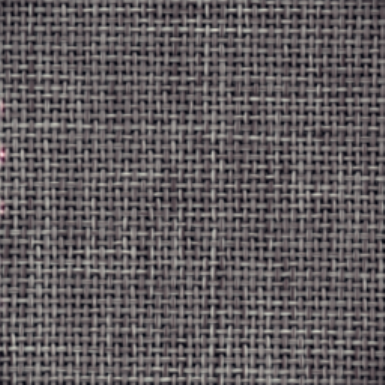} &
                \includegraphics*[width=0.21\columnwidth]{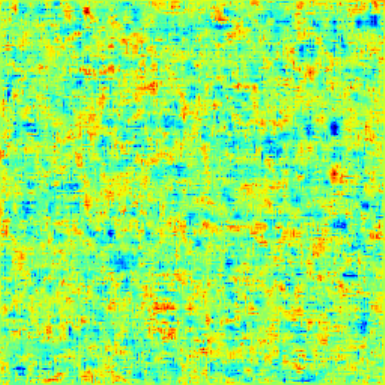} & &
    
                \includegraphics*[width=0.21\columnwidth]{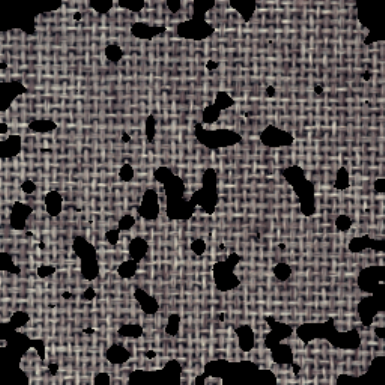} &
                \includegraphics*[width=0.21\columnwidth]{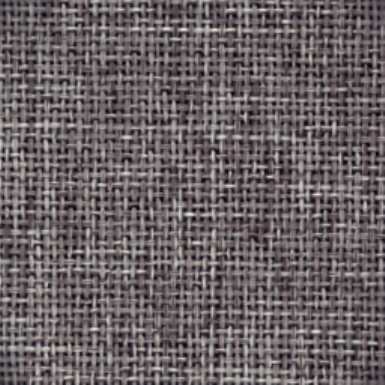} &
                \includegraphics*[width=0.21\columnwidth]{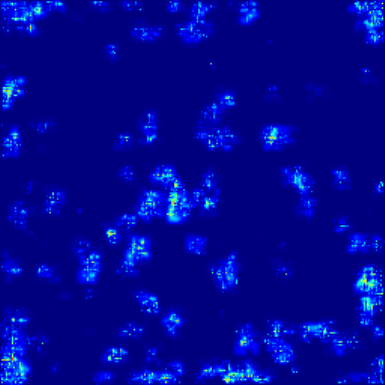} & &

                \includegraphics*[width=0.21\columnwidth]{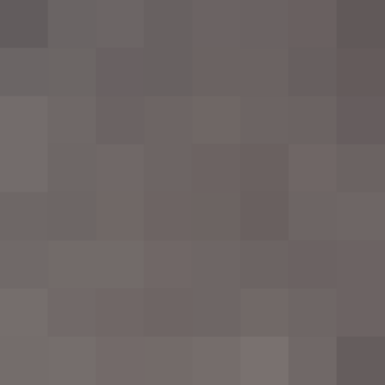} &
                \includegraphics*[width=0.21\columnwidth]{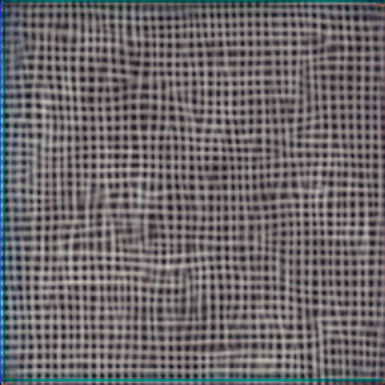} &
                \includegraphics*[width=0.21\columnwidth]{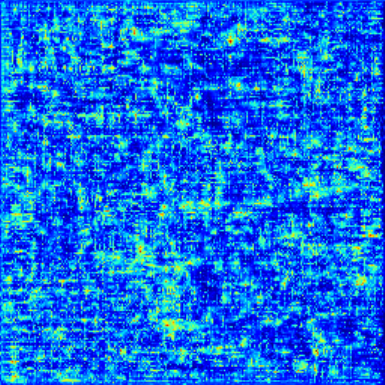} & &
    
                \includegraphics*[width=0.21\columnwidth]{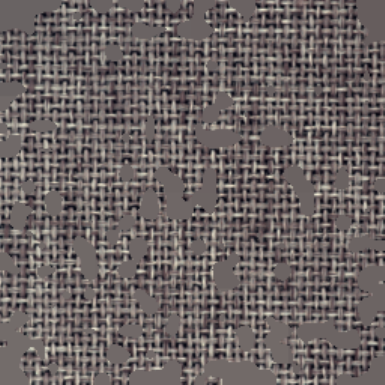} &
                \includegraphics*[width=0.21\columnwidth]{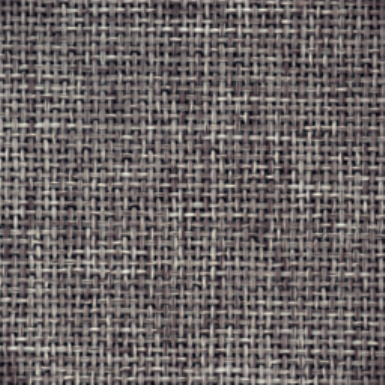} &
                \includegraphics*[width=0.21\columnwidth]{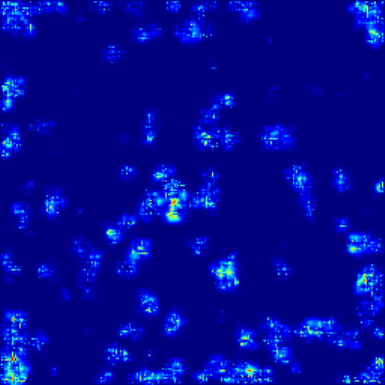} \\
                \vspace{-1.5cm} \\
    
                \rotatebox[origin=c]{90}{\qquad{}\qquad{}\qquad{}\quad{}Defective} & &
                \includegraphics*[width=0.21\columnwidth]{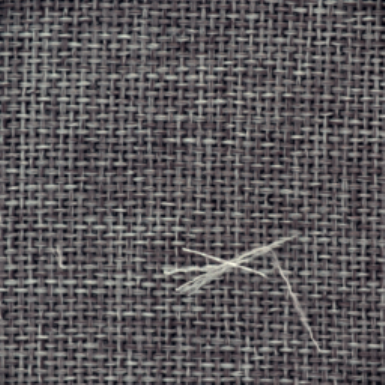} & &
                
                \includegraphics*[width=0.21\columnwidth]{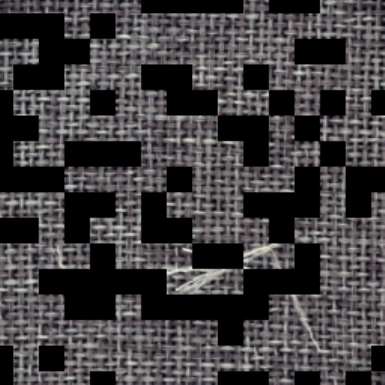} &
                \includegraphics*[width=0.21\columnwidth]{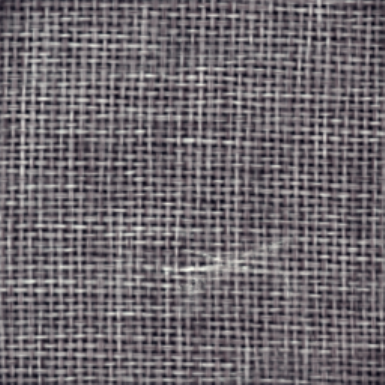} &
                \includegraphics*[width=0.21\columnwidth]{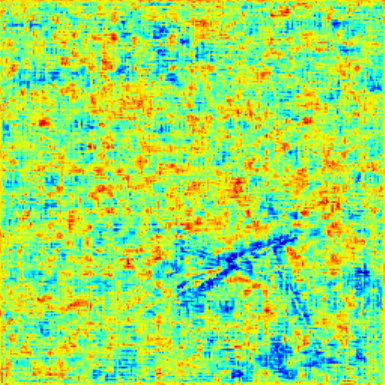} & &
    
                \includegraphics*[width=0.21\columnwidth]{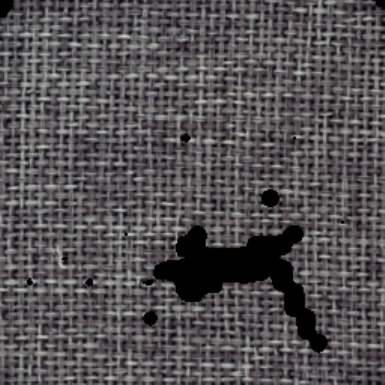} &
                \includegraphics*[width=0.21\columnwidth]{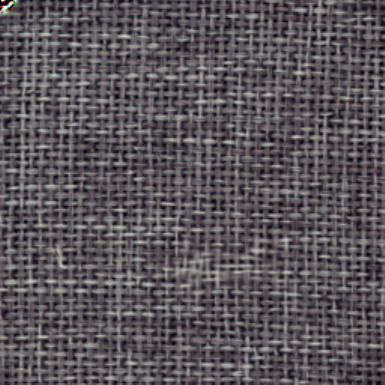} &
                \includegraphics*[width=0.21\columnwidth]{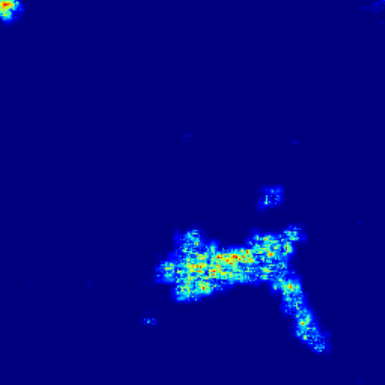} & &

                \includegraphics*[width=0.21\columnwidth]{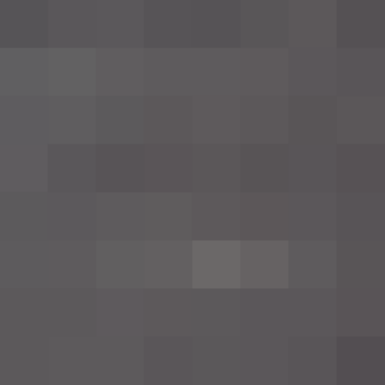} &
                \includegraphics*[width=0.21\columnwidth]{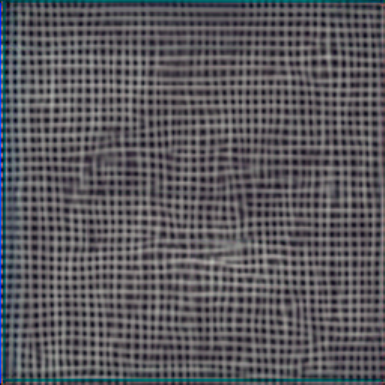} &
                \includegraphics*[width=0.21\columnwidth]{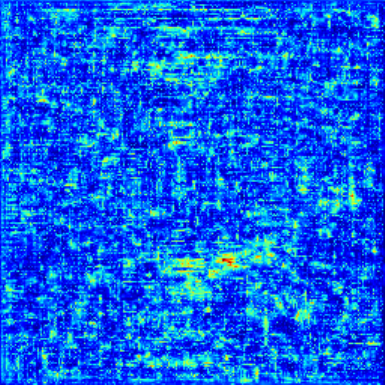} & &
    
                \includegraphics*[width=0.21\columnwidth]{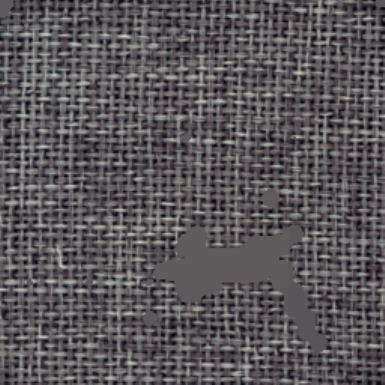} &
                \includegraphics*[width=0.21\columnwidth]{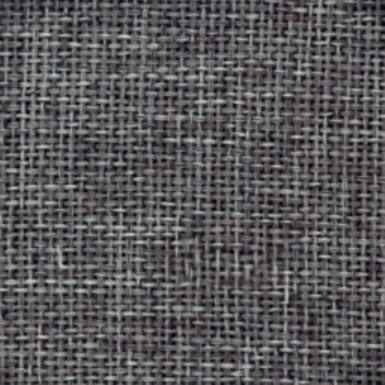} &
                \includegraphics*[width=0.21\columnwidth]{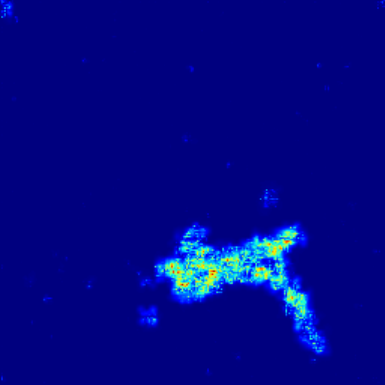} \\
                \vspace{-1.5cm} \\

            \hline
                \vspace{-0.25cm} \\
                \rotatebox[origin=c]{90}{\qquad{}\qquad{}\qquad{}\quad{}Normal} & &
                \includegraphics*[width=0.21\columnwidth]{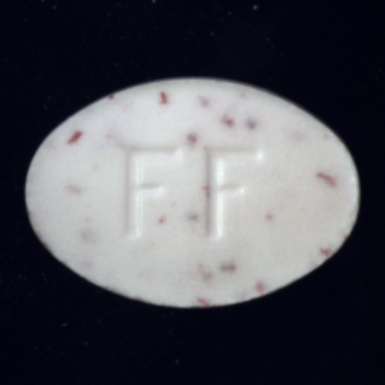} & &
                
                \includegraphics*[width=0.21\columnwidth]{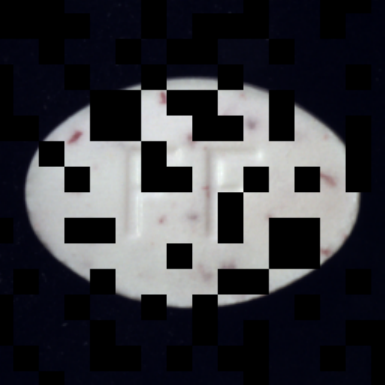} &
                \includegraphics*[width=0.21\columnwidth]{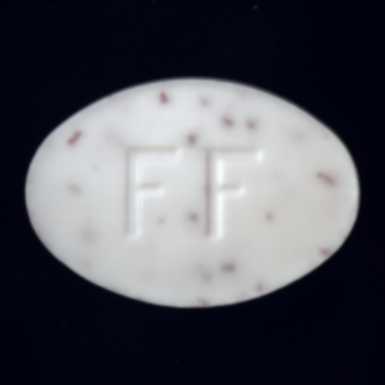} & 
                \includegraphics*[width=0.21\columnwidth]{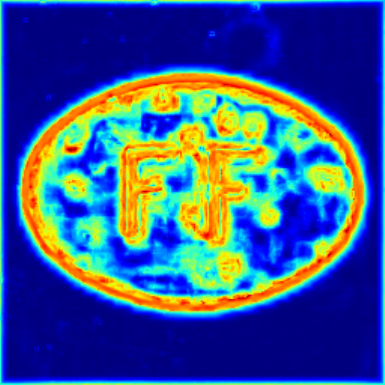} & &
    
                \includegraphics*[width=0.21\columnwidth]{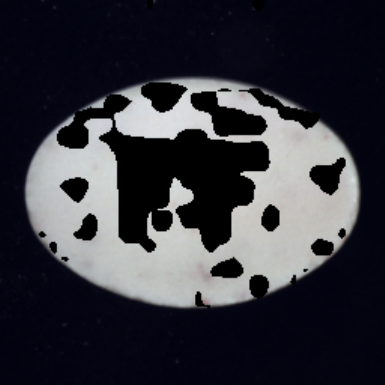} &
                \includegraphics*[width=0.21\columnwidth]{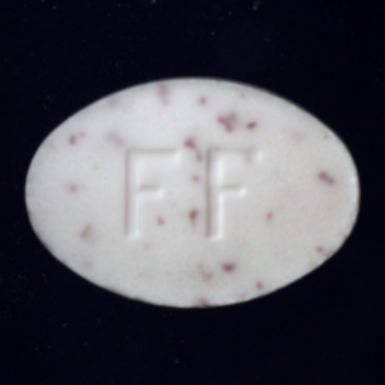} & 
                \includegraphics*[width=0.21\columnwidth]{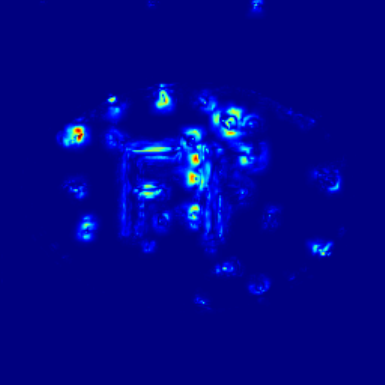} & &

                \includegraphics*[width=0.21\columnwidth]{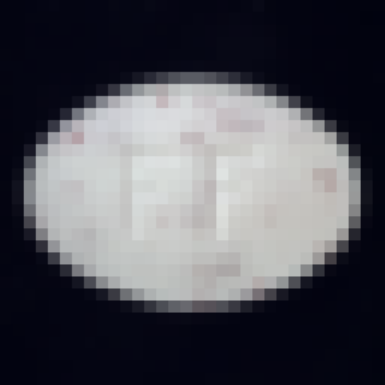} &
                \includegraphics*[width=0.21\columnwidth]{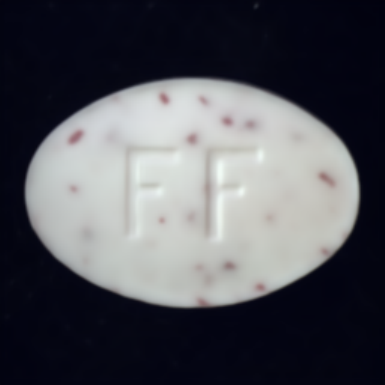} & 
                \includegraphics*[width=0.21\columnwidth]{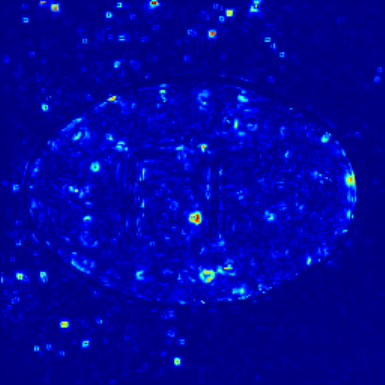} & &
    
                \includegraphics*[width=0.21\columnwidth]{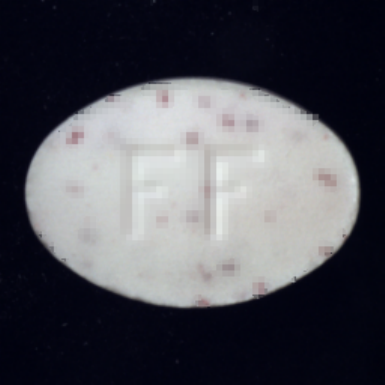} &
                \includegraphics*[width=0.21\columnwidth]{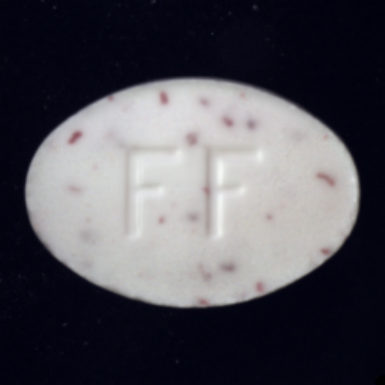} &
                \includegraphics*[width=0.21\columnwidth]{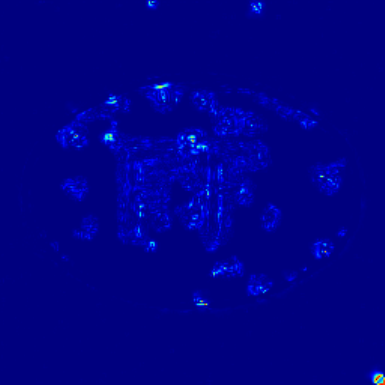} \\
                \vspace{-1.5cm} \\
    
                \rotatebox[origin=c]{90}{\qquad{}\qquad{}\qquad{}\quad{}Defective} & &
                \includegraphics*[width=0.21\columnwidth]{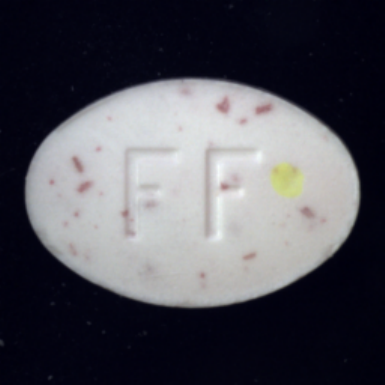} & &
                
                \includegraphics*[width=0.21\columnwidth]{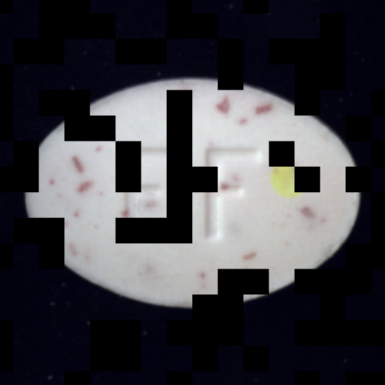} &
                \includegraphics*[width=0.21\columnwidth]{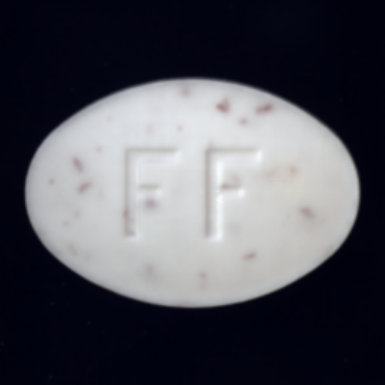} & 
                \includegraphics*[width=0.21\columnwidth]{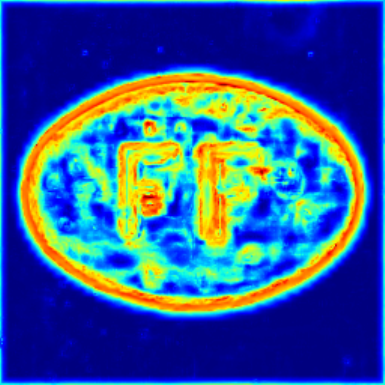} & &
    
                \includegraphics*[width=0.21\columnwidth]{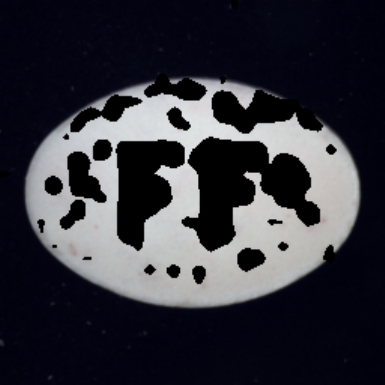} &
                \includegraphics*[width=0.21\columnwidth]{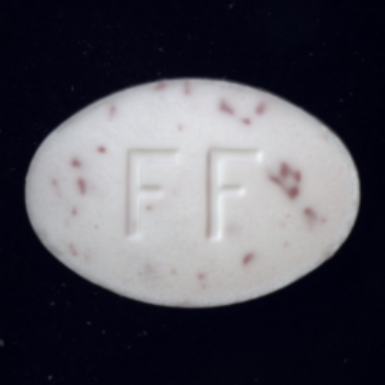} & 
                \includegraphics*[width=0.21\columnwidth]{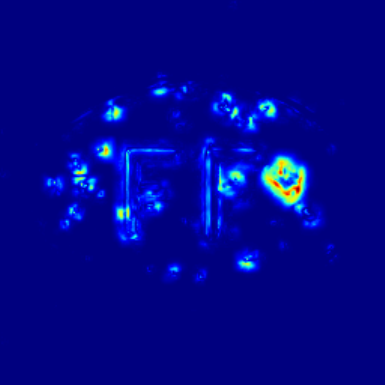} & &

                \includegraphics*[width=0.21\columnwidth]{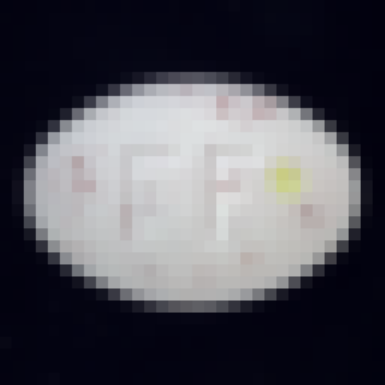} &
                \includegraphics*[width=0.21\columnwidth]{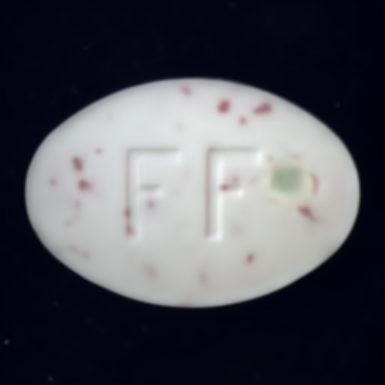} & 
                \includegraphics*[width=0.21\columnwidth]{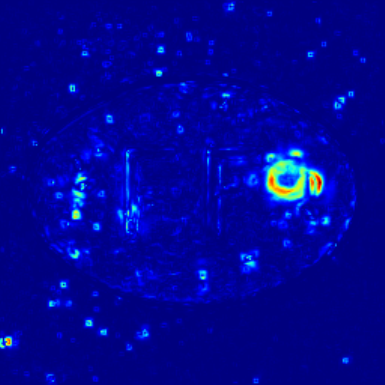} & &
    
                \includegraphics*[width=0.21\columnwidth]{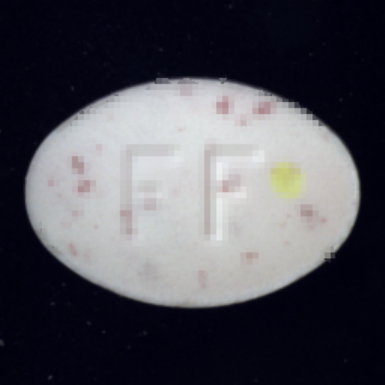} &
                \includegraphics*[width=0.21\columnwidth]{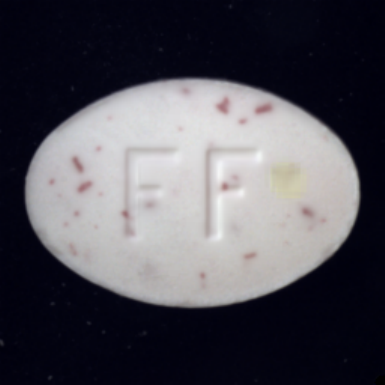} &
                \includegraphics*[width=0.21\columnwidth]{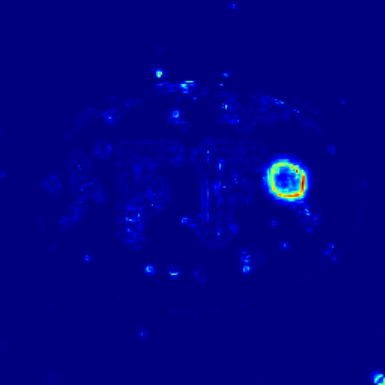}\\
                \vspace{-1.5cm} \\

        \end{tabular}
    }
    \vspace{-0.2cm}
    \caption{Visual comparison when the input corruption methods, including masking and mosaicing, vary. To visualize the results of RIAD~\cite{RIAD_Vitzan_PR21}, we have implemented and tested it for each subtask. The RIAD~\cite{RIAD_Vitzan_PR21} results show just one masking case among multiple disjoint masks and a cumulated error map for multiple inferences. They show large edge errors overall. For the red spot pattern at the pill, EAR$_{\mbox{\textit{w/o obf}}}$ shows a mistake in inpainting, and EAR$_{\mbox{\textit{w/o attn}}}$ produces scattered errors all over the region. EAR shows the accurate reconstruction of normal patterns by saliency masking and hint-providing by visual obfuscation. Best viewed in color.}
    \label{fig:vis_comparison}
    \vspace{-0.3cm}
\end{figure*}

\textbf{Implementation details.}
We simply inherit a well-known U-Net~\cite{Unet_Olaf_MICCAI15}-like structure as a reconstruction model for experiments.
Specifically, we construct an U-Net as in RIAD~\cite{RIAD_Vitzan_PR21}.
The reconstruction model is structured with five convolutional blocks for the encoder and decoder respectively, and the $i$-th layer in the encoder is concatenated with $(5$$-i)$-th layer in the decoder.
We repeat `convolution $\rightarrow$ batch normalization $\rightarrow$ leaky ReLU activation' three times for the encoder, and `upsampling $\rightarrow$ convolution $\rightarrow$ batch normalization $\rightarrow$ leaky ReLU activation'  three times for the decoder.
Note that, the stride is set to 2 in the third layer of each encoder block for spatial downscaling.
Also, upsampling with scaling factor 2 and the nearest interpolation is applied in the first layer of each decoder block.

To activate EAR, a pre-trained attention model is required.
Many variants of pre-trained ViT are publicly available.
We adopt one of the state-of-the-art models, specifically ViT-S/8.
It is provided in the official GitHub repository\footnote{https://github.com/facebookresearch/dino} published by Caron et al.~\cite{DINO_Caron_ICCV21}.
This ViT is known to have the ability to emphasize class-specific spatial features that lead to unsupervised object segmentation.
We leverage this property to emphasize saliency regions within a given image and cut them out for inpainting.

\textbf{Mosaic scale estimation.}
Our work includes an optimal mosaic scale estimation process. 
To construct a mosaic scale estimation model, we need the ground truth of optimal mosaic scale $m^{*}$ for each product.
We have initially found them in the grid search manner.
The results of finding the initial ground truth are given in Figure~\ref{fig:opt_mosaic}, and the linear regression model for mosaic scale estimation is shown in Figure~\ref{fig:estimation}.
A summary of mosaic scale estimation for each product is also given in Table~\ref{table:opt_mosaic}.
As can be seen in Table~\ref{table:opt_mosaic}, $m^{*}$ and $\hat{m}$ match in most objects.
However, there are mismatches for the textures.
Accordingly, we could observe that the AD performance for them using $\hat{m}$ is in general clearly lower than using $m^{*}$.
The experiments will be conducted with $\hat{m}$ to train an UAD model for each subtask.
The UAD results from $m^{*}$ will also be shown for comparison.

\textbf{Training conditions.}
We perform hyperparameter tuning in all UAD experiments for fair comparison of each model in the best performance condition.
The tuned hyperparameters are: 
1) kernel size 
2) learning rate 
3) scheduling method of learning rate. 
As learning rate scheduling methods, we used fixed learning rates, learning rate warm-up~\cite{Warm_Goyal_arXiv17}, and SGDR~\cite{SGDR_Ilya_ICLR17}.
The values used as hyperparameters are summarized in Table~\ref{table:hyperparameter}.

\textbf{Evaluation metric.}
To evaluate the performance of UAD experiments, we use the area under the receiver operating characteristic curve (AUROC)~\cite{AUROC_Tom_PRL06}.
The AUROC is measured based on the anomaly scores for each normal and defective sample within the test set.
For anomaly scoring, we adopt the maximum value of MSGMS between the input $I$ and reconstruction-by-inpainting result $\hat{I}$ of the UAD model which is capable of detecting various sizes of defects~\cite{RIAD_Vitzan_PR21}.
When the MSGMSs of the UAD model for the unseen anomalous patterns are relatively larger compared to the normal pattern cases, AUROC will be close to 1.

\begin{table*}[t]
    \centering
    \scriptsize
    \caption{Summary of the AUROC for the MVTec AD dataset~\cite{MVTec_Bergmann_CVPR19}. NNs are structured with simple well-known reconstruction backbones AE and U-Net~\cite{Unet_Olaf_MICCAI15}. For EAR, AUROCs are shown for two cases of $\hat{m}$ and $m^{*}$, in $\hat{m}$  ($m^{*}$) form. Abbreviations of attention module, discriminator, and memory module are `Att', `Dis', and `Mem' respectively.}
    \vspace{-0.2cm}
    \setlength\tabcolsep{4pt}
        \begin{tabular}{l || *{4}{P{1.15cm}} | *{3}{P{1.15cm}} | *{1}{P{2.2cm}}}
        \hline
            \multirow{2}{*}{Model} & MS-CAM & GANomaly & SCADN & MemAE & U-Net & DAAD & RIAD & EAR \\
            & \cite{MSCAM_Li_Sensors22} & \cite{GANomaly_Akcay_ACCV18} & \cite{SCADN_Yan_AAAI21} & \cite{MemAE_Gong_ICCV19} & \cite{Unet_Olaf_MICCAI15} & \cite{BWMem_Hou_ICCV21} & \cite{RIAD_Vitzan_PR21} & (ours) \\
        \hline
            Backbone & AE & AE & AE & AE & U-Net & U-Net & U-Net & U-Net \\
        \hline
            Additional & \multirow{2}{*}{Att} & \multirow{2}{*}{Dis} & \multirow{2}{*}{Dis} & \multirow{2}{*}{Mem} & \multirow{2}{*}{-} & Dis \& & \multirow{2}{*}{-} & \multirow{2}{*}{-} \\
            Module & & & & & & Mem & & \\
        \hline
        \hline
            Bottle & 0.940 & 0.892 & 0.957 & 0.930 & 0.863 & 0.976 & \textbf{0.999} & 0.997 (0.997) \\
            Cable & \textbf{0.880} & 0.732 & 0.856 & 0.785 & 0.636 & 0.844 & 0.819 & 0.853 (0.871) \\
            Capsule & 0.850 & 0.708 & 0.765 & 0.735 & 0.673 & 0.767 & \textbf{0.884} & 0.870 (0.870) \\
            Carpet & \textbf{0.910} & 0.842 & 0.504 & 0.386 & 0.774 & 0.866 & 0.842 & 0.850 (0.899) \\
            Grid & 0.940 & 0.743 & 0.983 & 0.805 & 0.857 & 0.957 & \textbf{0.996} & 0.952 (0.959) \\
            Hazelnut & 0.950 & 0.794 & 0.833 & 0.769 & 0.996 & 0.921 & 0.833 & \textbf{0.997 (0.997)} \\
            Leather & 0.950 & 0.792 & 0.659 & 0.423 & 0.870 & 0.862 & \textbf{1.000} & \textbf{1.000 (1.000)} \\
            Metal nut & 0.690 & 0.745 & 0.624 & 0.654 & 0.676 & 0.758 & \textbf{0.885} & 0.856 (0.876) \\
            Pill & 0.890 & 0.757 & 0.814 & 0.717 & 0.781 & 0.900 & 0.838 & \textbf{0.922 (0.922)} \\
            Screw & \textbf{1.000} & 0.699 & 0.831 & 0.257 & \textbf{1.000} & 0.987 & 0.845 & 0.779 (0.886) \\
            Tile & 0.800 & 0.785 & 0.792 & 0.718 & 0.964 & 0.882 & \textbf{0.987} & 0.918 (0.965) \\
            Toothbrush & \textbf{1.000} & 0.700 & 0.891 & 0.967 & 0.811 & 0.992 & \textbf{1.000} & \textbf{1.000 (1.000)} \\
            Transistor & 0.880 & 0.746 & 0.863 & 0.791 & 0.674 & 0.876 & 0.909 & \textbf{0.947 (0.947)} \\
            Wood & 0.940 & 0.653 & 0.968 & 0.954 & 0.958 & 0.982 & 0.930 & \textbf{0.946 (0.985)} \\
            Zipper & 0.910 & 0.834 & 0.846 & 0.710 & 0.750 & 0.859 & \textbf{0.981} & 0.949 (0.955) \\
        \hline
        \hline
            Average & 0.902 & 0.761 & 0.812 & 0.707 & 0.819 & 0.895 & 0.917 & \textbf{0.922 (0.942)} \\
        \hline
        \end{tabular}
    \label{table:perform_mvtec}
    \vspace{-0.4cm}
\end{table*}

\subsection{Visual comparison of reconstruction}
We report visual comparisons of reconstruction results to see whether the reconstruction of normal patterns is accurate when the proposed method EAR is applied.
First, as can be seen in Figure~\ref{fig:smore_ear}, we check the effect of disabling either of the important design components: saliency masking and visual obfuscation by mosaicing.
EAR$_{\mbox{\textit{w/o obf}}}$ disables visual obfuscation for hint-providing.
That is, EAR$_{\mbox{\textit{w/o obf}}}$ empties saliency region without any hint.
EAR$_{\mbox{\textit{w/o attn}}}$ does not utilize the ImageNet~\cite{ImageNet_Deng_CVPR09} pre-trained DINO-ViT~\cite{DINO_Caron_ICCV21} to cut out suspected anomalous regions.
Thus, EAR$_{\mbox{\textit{w/o attn}}}$ reconstructs the whole image that is obfuscated.
The results for each model are shown for the best hyperparameter conditions. 
EAR$_{\mbox{\textit{w/o obf}}}$ shows an inpainting mistake on the background region marked with a red box, and EAR$_{\mbox{\textit{w/o attn}}}$ struggles reconstructing the normal region within the defective sample.
The full model, EAR, accurately reconstructs normal regions within a defective sample, marked in the yellow box. 
In addition, red-boxed anomalous regions are successfully transformed into a normal form without inpainting mistakes. 
Those reconstruction results confirm that both saliency masking and mosaic obfuscation for hint-providing play an essential role in achieving \textit{contained generalization ability}~\cite{LAMP_Park_arXiv23} by complementing each other.

In Figure~\ref{fig:vis_comparison}, we present visual comparisons of the EAR variants with RIAD~\cite{RIAD_Vitzan_PR21}.
RIAD~\cite{RIAD_Vitzan_PR21} features reconstruction by inpainting with multiple disjoint random masks and provides a cumulated error map.
The reconstruction results of RIAD~\cite{RIAD_Vitzan_PR21} are just one case from multiple masking.
The overall edge error, MSGMS, is large due to a lot of random patch masks being located with the edge of the object in RIAD~\cite{RIAD_Vitzan_PR21} case.
EAR$_{\mbox{\textit{w/o obf}}}$ case shows the inaccurate reconstruction of the normal region by a large binary mask. 
Especially in a defective pill case, binary masking causes confusion as to whether the empty space should be filled with a red dot pattern or a white color.
EAR$_{\mbox{\textit{w/o attn}}}$ shows scattered minute errors all over the region are produced because of the spatial discontinuity of the input image due to the mosaic. 
In contrast, EAR accurately reconstructs normal patterns by leveraging the hint-providing strategy from mosaic obfuscation; specifically, the logo and digit printings of the capsule. 
It successfully transforms the scratched white digit printing `500' into a normal form.

\subsection{Anomaly detection performance}
We train EAR with anomaly-free samples.
Then, the AUROC is measured for each subtask with an MSGMS-based anomaly scoring method.
The measured performance is summarized in Table~\ref{table:perform_mvtec}.
As our study proposes a strategy to maximize the performance without changing the NN structure, we compare the performance with recent studies that use NNs of the same or similar scale.

EAR achieves the best performance in hazelnut, pill, and transistor cases compared to other models.
The common characteristic of defective samples in these subtasks is surface damage which can be recovered into normal form by EAR.
In cases of capsules, screws, and zippers that show sophisticated features, AUROC is relatively low compared to the highest performance.
This is because the detailed pattern alignment of screw thread or the zipper teeth by reconstruction may be slightly missed due to saliency masking and visual obfuscation in the suspected anomalous regions.
In the other cases that show AUROC close to 1, EAR also achieves similar cutting-edge performance with a very slight margin.

In conclusion, EAR achieves AUROCs of 0.922 and 0.942 by utilizing $\hat{m}$ and $m^{*}$ respectively.
The performance 0.922 with $\hat{m}$ indicates that EAR with mosaic scale estimation enhances the UAD performance compared to prior state-of-the-art models.
When the mosaic scale tuning with grid search is conducted for each subtask to use the true optimal mosaic scale, we can further enhance the UAD performance.

\subsection{Training and inference speed}
The measured processing time is summarized in Table~\ref{table:time}.
RIAD~\cite{RIAD_Vitzan_PR21} takes the longest time for both training and inference due to the multiple masking strategy.
For reference, the number of masks used in RIAD~\cite{RIAD_Vitzan_PR21} is set to 12 as suggested in the original paper.
In Table~\ref{table:time}, EAR$_{\mbox{\textit{w/o attn}}}$ shows the fastest speed because it does not involve generating saliency maps through a pre-trained attention model, ImageNet~\cite{ImageNet_Deng_CVPR09} pre-trained DINO-ViT~\cite{DINO_Caron_ICCV21}.
EAR$_{\mbox{\textit{w/o obf}}}$ and EAR are somewhat slower than EAR$_{\mbox{\textit{w/o attn}}}$ because they generate saliency maps via a pre-trained attention model.
They show 2.35 $\times$ and 1.86 $\times$ faster inference than RIAD~\cite{RIAD_Vitzan_PR21}, respectively.
EAR shows an inference speed fast enough for real-time processing with the highest UAD performance.

\begin{table}[t]\centering
    \footnotesize
    \caption{Processing time for each training and inference. In inference, EAR$_{\mbox{\textit{w/o attn}}}$ is the fastest model, also EAR shows sufficiently fast inference speed of is 1.86 $\times$ faster than RIAD~\cite{RIAD_Vitzan_PR21}.}
    \vspace{-0.2cm}
    \setlength\tabcolsep{2pt}
        \begin{tabular}{l||r|r}
        \hline
            Model & Training (sec) & Inference (msec) \\
        \hline
        \hline
            RIAD~\cite{RIAD_Vitzan_PR21} & 35,478 & 366 \\
            EAR$_{\mbox{\textit{w/o obf}}}$ & 3,084 & 156 \\
            EAR$_{\mbox{\textit{w/o attn}}}$ & \textbf{3,078} & \textbf{37} \\
            EAR & 3,109 & 197 \\
        \hline
        \end{tabular}
    \label{table:time}
    \vspace{-0.5cm}
\end{table}

\subsection{Ablation study}
We have conducted an ablation study to see how deterministic saliency masking and obfustication by mosaicing for hint-providing affect the UAD performance.
In addition, we also check the effect of applying the knowledge distillation (KD) method, part of SQUID~\cite{SQUID_Xiang_CVPR23}, which uses two of the same NNs as teacher and student, respectively, during the training stage. 

The results of the ablation study are summarized in Table~\ref{table:ablation}.
The 2nd column shows the cases of EAR$_{\mbox{\textit{w/o obf}}}$.
It appears to be difficult to achieve a high performance because binary masking empties all the information in the suspected defective regions, causing inaccurate reconstruction on both normal and anomalous patterns.
On the other hand, EAR$_{\mbox{\textit{w/o attn}}}$, shown in the 3rd column, confirms that the obfustication by mosaicing achieves better UAD performance compared to RIAD~\cite{RIAD_Vitzan_PR21} and EAR$_{\mbox{\textit{w/o obf}}}$ because of the relatively accurate reconstruction of normal patterns, reducing false positives. 
Also refer to the visual comparisons of the above cases in Figure~\ref{fig:vis_comparison}.

We confirm that the UAD performance is further improved when mosaic is used for hint-providing as shown in the last column, which is the case of EAR.
Note that, the full model of EAR exploits both hint-providing and pre-trained attention-based saliency masking.

When the KD strategy from SQUID~\cite{SQUID_Xiang_CVPR23} is additionally applied (the 5th column in Table~\ref{table:ablation}), there is almost no change in the performance.
Referring to the expensive training cost of KD due to the use of two identical NNs (for each teacher and student), we do not see any advantage of additionally employing KD for EAR.

\begin{table}[t]
    \centering
    \scriptsize
    \caption{Summary of the ablation study. Except for RIAD~\cite{RIAD_Vitzan_PR21}, all the other masking cases employ a pre-trained attention-based deterministic single saliency masking. The rightmost two columns report the performance of visual defect obfuscation methods when using $\hat{m}$ and $m^{*}$, in $\hat{m}$  ($m^{*}$) form.}
    \vspace{-0.2cm}
    \setlength\tabcolsep{3pt}
        \begin{tabular}{l|| c | ccc | c}
            \hline
                \multirow{2}{*}{Model} & RIAD & \multicolumn{3}{c|}{\multirow{2}{*}{Ablations}} & EAR \\
                & \cite{RIAD_Vitzan_PR21} & & & & (ours) \\
            \hline
            \hline
                Masking & \cmark (multi) & \cmark & & \cmark & \cmark \\
            \hline
                Hint & & & \cmark & \cmark & \cmark \\
            \hline
                KD~\cite{SQUID_Xiang_CVPR23} & & & & \cmark & \\
            \hline
            \hline
                Bottle & 0.999 & 0.995 & \textbf{1.000}  & 0.994 (0.995) & 0.997 (0.997) \\
                Cable & 0.819 & 0.795 & \textbf{0.888}  & 0.851 (0.855) & 0.853 (0.871) \\
                Capsule & 0.884 & 0.784 & \textbf{0.918}  & 0.869 (0.869) & 0.870 (0.870) \\
                Carpet & 0.842 & 0.848 & 0.718  & 0.846 (0.880) & \textbf{0.850 (0.899)} \\
                Grid & \textbf{0.996} & 0.969 & 0.963  & 0.976 (0.976) & 0.952 (0.959) \\
                Hazelnut & 0.833 & 0.986 & 0.996  & 0.992 (0.996) & \textbf{0.997 (0.997)} \\
                Leather & \textbf{1.000} & \textbf{1.000} & \textbf{1.000} & \textbf{1.000 (1.000)} & \textbf{1.000 (1.000)} \\
                Metal nut & \textbf{0.885} & 0.832 & 0.841  & 0.868 (0.868) & 0.856 (0.876) \\
                Pill & 0.838 & 0.738 & 0.867  & 0.870 (0.873) & \textbf{0.922 (0.922)} \\
                Screw & \textbf{0.845} & 0.800 & 0.825 & 0.776 (0.854) & 0.779 (0.886) \\
                Tile & \textbf{0.987} & 0.928 & 0.939  & 0.956 (0.956) & 0.918 (0.965) \\
                Toothbrush & \textbf{1.000} & 0.994 & \textbf{1.000} & \textbf{1.000 (1.000)} & \textbf{1.000 (1.000)} \\
                Transistor & 0.909 & 0.891 & 0.943  & 0.895 (0.933) & \textbf{0.947 (0.947)} \\
                Wood & 0.930 & 0.904 & 0.945  & \textbf{0.986 (0.995)} & 0.946 (0.985) \\
                Zipper & \textbf{0.981} & 0.900 & 0.963  & 0.951 (0.961) & 0.949 (0.955) \\
            \hline
            \hline
                Average & 0.917 & 0.891 & 0.920  & 0.922 (0.934) & \textbf{0.922 (0.942)} \\
            \hline
        \end{tabular}
    \label{table:ablation}
    \vspace{-0.5cm}
\end{table}

\vspace{-0.1cm}
\section{Conclusion}
In this study, we propose a novel self-supervised learning strategy, EAR, to enhance the UAD-purposed reconstruction-by-inpainting model.
We have effectively exploited the ImageNet~\cite{ImageNet_Deng_CVPR09} pre-trained DINO-ViT~\cite{DINO_Caron_ICCV21} to generate a deterministic single saliency mask to cut out suspected anomalous regions.
EAR also provides the best possible hint for reconstruction by visual obfuscation with the proper mosaic scale estimation. 
EAR not only serves the reliability of resulting the output via deterministic masking and hint-providing strategy but also achieves fast inference via single masking.
Moreover, the UAD performance is enhanced because hint-providing strategy promotes the accurate reconstruction of normal patterns and effective translation of anomalous patterns into a normal form.

Our approach proposed in this study is distinguished from others by enhancing the UAD performance with computational efficiency.
Thus, we suggest EAR for various manufacturing industries as a practically deployable solution.

\vspace{-0.1cm}
\section*{Acknowledgment}
\vspace{-0.15cm}
This research was supported by SK Planet Co., Ltd. and Computer Vision Lab at Sungkyunkwan University.

{ \small
\bibliographystyle{ieee_fullname}
\bibliography{egbib}
}

\end{document}